\documentclass[journal]{IEEEtran}
\usepackage{float}
\usepackage{color}
\usepackage{comment}
\usepackage{colortbl}  
\usepackage{xcolor}
\usepackage{epsfig}
\usepackage{graphicx}
\usepackage{bbding}
\usepackage{array, caption, threeparttable}
\usepackage{amsmath}
\usepackage{amssymb}
\usepackage{cite}
\usepackage[numbers,sort&compress]{natbib}
\usepackage{comment}
\usepackage{multirow}
\usepackage{placeins} 
\usepackage{hyperref}
\hypersetup{hidelinks}
\usepackage{makecell} 
\usepackage{booktabs}
\usepackage{array, caption, threeparttable}
\usepackage{caption}
\usepackage{subcaption}
\usepackage[ruled]{algorithm2e}
\usepackage{listings}
\usepackage{color}
\usepackage{mathtools}

\usepackage{todonotes}
\usepackage{gensymb}
\usepackage{setspace}

\usepackage{algorithmic}

\usepackage[normalem]{ulem}
\usepackage{cancel}

\begin{document}

\title{\LARGE \bf CardiacMamba: A Multimodal RGB-RF Fusion Framework with State Space Models for Remote Physiological Measurement}

\author{Zheng Wu, Yiping Xie, Bo Zhao, Jiguang He, Fei Luo, Ning Deng, Zitong Yu
\thanks{This work was supported by National Natural Science Foundation of China (Project No. 62306061), and Guangdong Research Team for Communication and Sensing Integrated with Intelligent Computing (Project No. 2024KCXTD047). Zheng Wu and Yiping Xie are co-first authors and contribute equally. Corresponding author: Zitong Yu (email: zitong.yu@ieee.org).}
\thanks{Z. Wu, B. Zhao, J. He, F. Luo, N. Deng, and Z. Yu are with School of Computing and Information Technology, Great Bay University, and also with Dongguan Key Laboratory for Intelligence and Information Technology, Dongguan, 523000, China.}
\thanks{Y. Xie is with Computer Vision Institute, School of Computer Science \& Software Engineering, Shenzhen University, Shenzhen, 518060, and also with School of Computing and Information Technology, Great Bay University, Dongguan, 523000, China.}
}

\markboth{IEEE Transactions on Instrumentation and Measurement}%
{Shell \MakeLowercase{\textit{et al.}}: Bare Demo of IEEEtran.cls for IEEE Journals}

\maketitle


\begin{abstract}

Heart rate (HR) estimation via remote photoplethysmography (rPPG) offers a non-invasive solution for health monitoring. However, traditional single-modality approaches (RGB or Radio Frequency (RF)) face challenges in balancing robustness and accuracy due to lighting variations, motion artifacts, and skin tone bias. In this paper, we propose CardiacMamba, a multimodal RGB-RF fusion framework that leverages the complementary strengths of both modalities. It introduces the Temporal Difference Mamba Module (TDMM) to capture dynamic changes in RF signals using timing differences between frames, enhancing the extraction of local and global features. Additionally, CardiacMamba employs a Bidirectional SSM for cross-modal alignment and a Channel-wise Fast Fourier Transform (CFFT) to effectively capture and refine the frequency domain characteristics of RGB and RF signals, ultimately improving heart rate estimation accuracy and periodicity detection. Extensive experiments on the EquiPleth dataset demonstrate state-of-the-art performance, achieving marked improvements in accuracy and robustness. CardiacMamba significantly mitigates skin tone bias, reducing performance disparities across demographic groups, and maintains resilience under missing-modality scenarios. By addressing critical challenges in fairness, adaptability, and precision, the framework advances rPPG technology toward reliable real-world deployment in healthcare. The codes are available at: \href{https://github.com/WuZheng42/CardiacMamba}{https://github.com/WuZheng42/CardiacMamba}.


\end{abstract}


\IEEEpeerreviewmaketitle

\section{Introduction}

\IEEEPARstart{H}{eart} rate (HR) is a crucial physiological signal for assessing a person’s health and emotional state. Traditional HR measurement methods like Electrocardiography (ECG) and Photoplethysmography (PPG) require contact sensors, which can be uncomfortable and impractical for long-term monitoring. To address these limitations, remote photoplethysmography (rPPG) has emerged as a promising solution. rPPG uses facial videos to capture skin color changes caused by blood flow~\cite{ahad2021contactless,cheng2020remote,liu2024remote,hu2021eta,yu2021facial}, enabling non-contact HR estimation. This method is gaining attention due to its potential applications in various fields, such as healthcare, human-robot interaction, and driver monitoring systems. Additionally, rPPG offers advantages over traditional sensors by being non-invasive, convenient, and cost-effective, making it suitable for continuous health monitoring.

Non-contact physiological systems, such as those using cameras for remote photoplethysmography (rPPG), rely on the movement of the skin's RGB spectrum with the blood volume pulse (BVP), providing insights into human vital signs. The field of camera-based remote plethysmography has evolved significantly, with methods based on facial skin color changes linked to blood volume, using approaches like physics-based models~\cite{dehaan2013robust,verkruysse2008remote}, blind source separation~\cite{lewandowska2011measuring,poh2010advancements}, and deep learning based methods~\cite{yu2019remote}. However, environmental factors (e.g., illumination variations) severely degrade performance. For instance, under low-light conditions, darker skin tones further amplify estimation errors due to reduced light reflectance~\cite{nowara2018sparseppg}.

\begin{figure}
\centering
\begin{subfigure}{0.5\textwidth}
    \includegraphics[width=\linewidth, height=3cm]{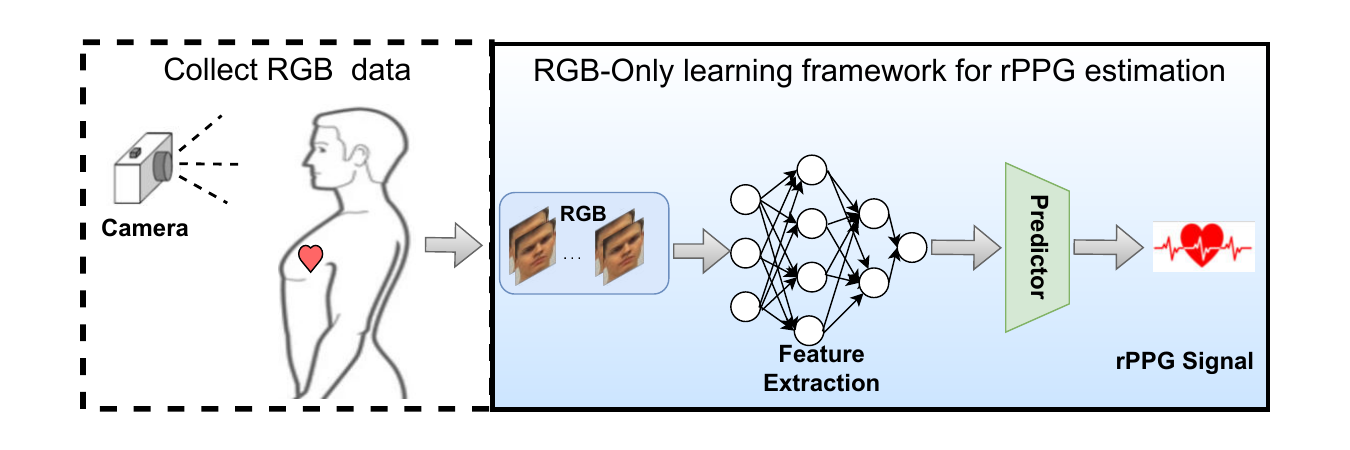}
    \vspace{-2.0em}
    \caption{RGB-only Methods}
    \label{fig:rgb}
\end{subfigure}
\hfill
\begin{subfigure}{0.5\textwidth}
    \centering
    \includegraphics[width=\linewidth, height=3cm]{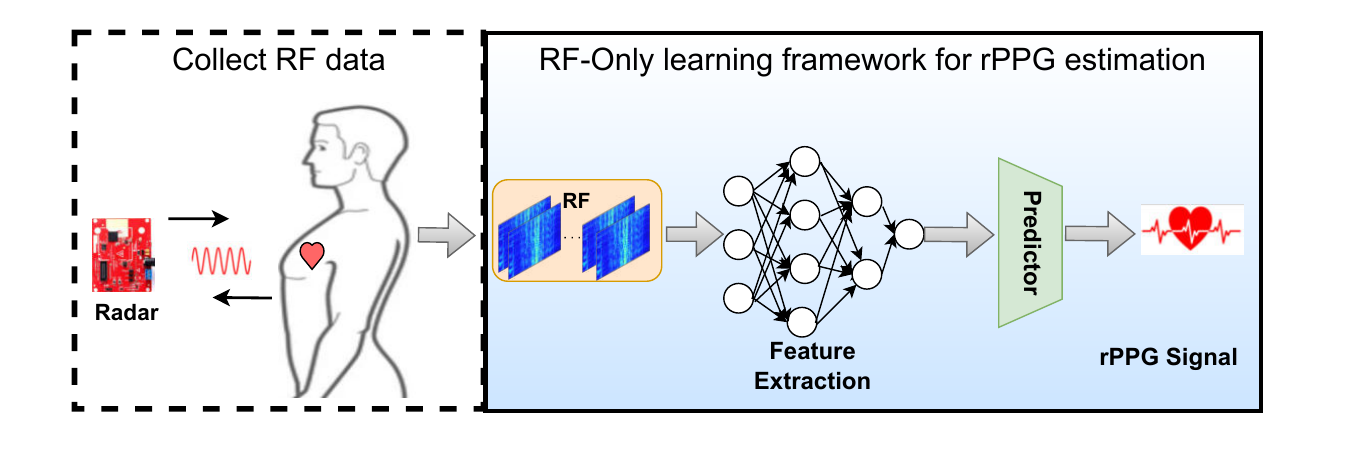}
    \vspace{-2.0em}
    \caption{RF-only Methods}
    \label{fig:rf}
\end{subfigure}
\hfill
\begin{subfigure}{0.5\textwidth}
    \centering
    \includegraphics[width=\linewidth, height=3cm]{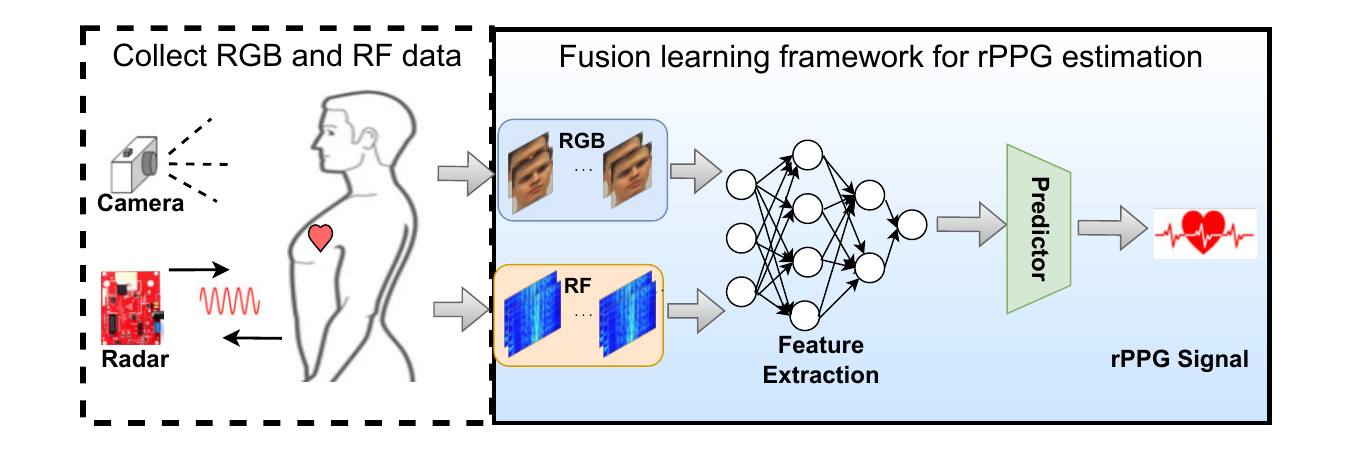}
    \vspace{-2.0em}
    \caption{RGB-RF fusion Methods}
    \label{fig:rgb_rf}
\end{subfigure}
\caption{Comparison of deep learning methods for rPPG learning. (a) RGB-only Method: Training with only RGB data collected by the camera. (b) RF-only Method: Training with only RF data collected by the radar. (c) Training with both RGB and RF data. }
\label{fig:all_cameras}
\vspace{-1.0em}
\end{figure}




To address the limitations of video-based methods, alternative non-contact solutions have emerged, offering potential advantages in overcoming these challenges. Radio Frequency (RF) sensors estimate heart rate by periodically emitting and receiving electromagnetic signals to measure the radial depth changes near the chest, which vibrates in response to the individual's vital cycle. Unlike video-based physiological monitoring methods that rely on RGB pixel intensity~\cite{park2019noncontact,zheng2020v2ifi}, RF sensors infer radial depth information through electromagnetic signal transmission and reception, inherently mitigating the influence of surrounding illumination. However, RF sensors have their own disadvantages, such as poor angular resolution that makes them susceptible to lateral motions, and the greater difficulty they pose for data acquisition compared to cameras. Consequently, most previous RF physiological approaches have depended upon learning-free methods~\cite{alizadeh2019remote, li2008random}. But in recent years, deep learning-based technologies have also emerged~\cite{wu2019person}.

The fusion of RGB and RF modalities presents a compelling avenue to synergize their complementary strengths: rPPG provides high spatial resolution for localized signal extraction, while RF ensures robustness against lighting variations and skin tone biases. However, current multimodal fusion strategies inadequately address three key gaps~\cite{vilesov2022blending}: (1) inefficient cross-modal feature alignment due to heterogeneous temporal and spatial characteristics, (2) insufficient exploitation of frequency-domain correlations for periodic HR signal enhancement, and (3) persistent fairness disparities in underrepresented demographic groups. To address the limitations of video-based methods, alternative non-contact solutions have emerged, offering potential advantages in overcoming these challenges. 

In response to the above issues, this paper proposes CardiacMamba, a multi-modal RGB-RF fusion framework based on dynamic feature enhancement and cross-modal bidirectional interaction using a State Space Model. The framework aims to significantly improve the performance and fairness of heart rate estimation and other tasks by introducing innovative module designs and frequency-domain fusion strategies. In the Dual-level Feature Extraction and Alignment stage, we introduce the Temporal Difference Mamba Module (TDMM) and the Bifurcated Diff-Conv Fusion (BDCF). These modules extract dynamic features through temporal differential convolutions, and, combined with the global modeling capability of the Mamba block, effectively enhance the rPPG features of the RGB modality and the temporal dynamic information of the RF modality. In the Bidirectional Feature Interaction stage, we innovatively introduce the Bidirectional State Space Model (Bi-SSM). This model enables the cooperative modeling of the RGB and RF modalities by sharing state transition matrices and input matrices. The mechanism integrates cross-modal information under a unified dynamic framework, preserving the unique semantics of each modality while enhancing global context awareness through bidirectional temporal modeling (both forward and backward). In the Bidirectional Feature Fusion stage, we propose the Channel-wise Fast Fourier Transform, which maps the RGB and RF features into the frequency domain for interaction. By utilizing learnable real and imaginary part interaction parameters, the model can adaptively enhance the frequency bands related to heart rate and suppress noise, with the final reconstruction back into time-domain features through inverse transformation. Experiments on the EquiPleth dataset demonstrate that CardiacMamba outperforms existing methods in scenarios involving skin tone bias and missing modalities. Our multi-modal fusion mechanism effectively alleviates the estimation bias caused by skin reflectance differences in traditional single-modal methods.

In summary, our contributions encompass:

\begin{itemize}
    \item We introduce Mamba into the framework for joint RGB and RF fusion-based rPPG estimation, enhancing the model’s ability to efficiently capture and integrate multi-modal features. By leveraging Mamba’s advanced structure, we improve the robustness and accuracy of rPPG estimation, enabling the model to effectively handle the diverse and complementary information from both RGB and RF modalities.
    
    \item We design the Temporal Difference Mamba Module (TDMM), which improves the standard Mamba framework by leveraging temporal differences between consecutive frames to capture dynamic changes in RF signals. By incorporating the Mamba block, this module enhances the extraction of both local and global temporal features, thereby significantly improving the model's performance in complex dynamic scenarios.

    \item We designed the Channel-wise Fast Fourier Transform (CFFT), which effectively captures the unique frequency-domain characteristics of RGB and RF modalities. This frequency-domain approach enhances the accuracy of heart rate estimation by revealing cross-modal frequency relationships and selectively enhancing or suppressing relevant components, thereby strengthening the model’s ability to detect heart rate periodicity.
    
    \item We have achieved state-of-the-art performance in handling missing modality scenarios and eliminating skin tone bias, demonstrating the advantages of our proposed model in dealing with incomplete data and cross-modal data biases. At the same time, we conducted a series of extensive ablation experiments to thoroughly evaluate the individual components of the model, validating the critical role each module plays in enhancing overall performance.

\end{itemize}

\begin{figure*}
\centering
\vspace{-1.5em}
\includegraphics[width=1\linewidth]{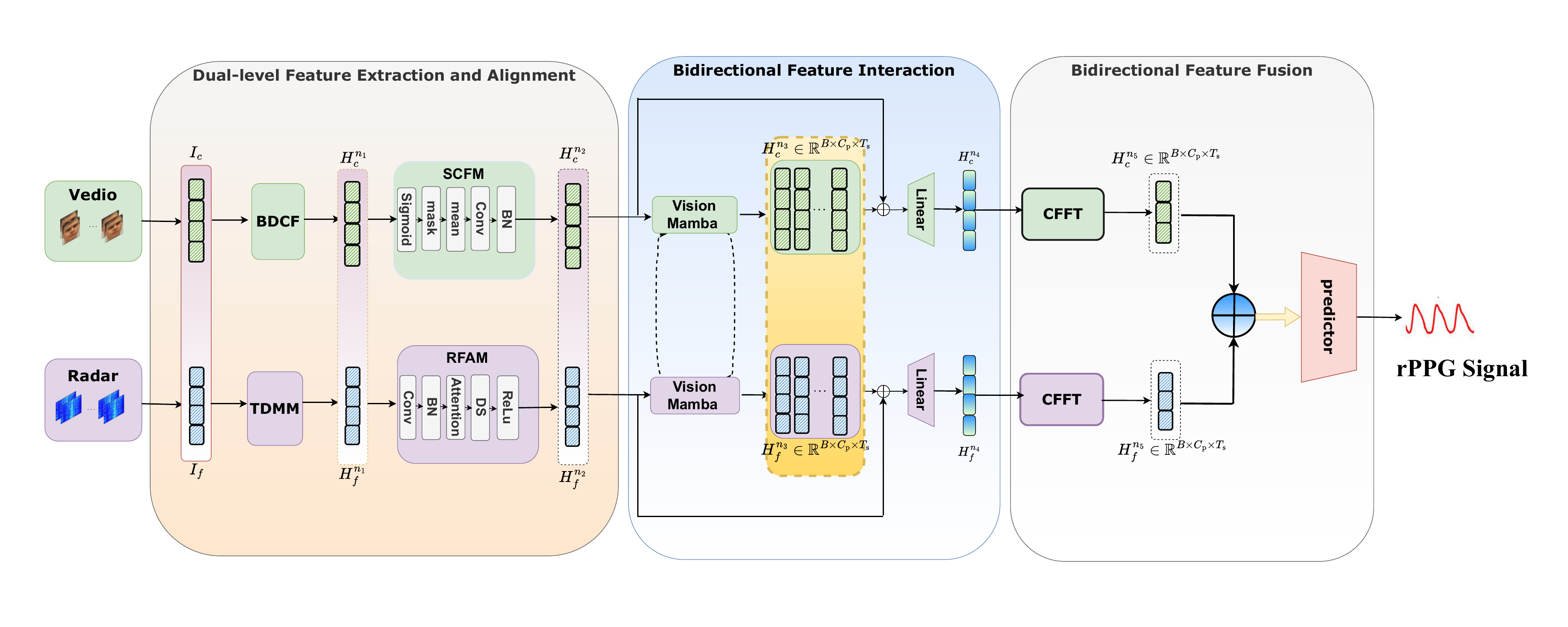}
\vspace{-3.0em}

  \caption{{The overall architecture of CardiacMamba. It consists of three stages: Dual-level Feature Extraction and Alignment, Bidirectional Feature Interaction, and Bidirectional Feature Fusion. 
}}
  \label{fig:network}
  \vspace{-1.0em}
\end{figure*}

\vspace{-1.0em}
\section{Related Work}
\label{sec:relatedwork}

\subsection{RGB Video-Based Methods}

Camera-based remote photoplethysmography (rPPG) technology has undergone significant development over the past few decades. In video-based remote physiological monitoring, RGB cameras can be used to remotely reconstruct human physiological information, particularly in the facial region, because the skin’s reflectance spectrum changes with physiological movements such as blood pulses. Initially, traditional rPPG methods mainly relied on signal processing techniques to analyze periodic signals from facial regions. These methods typically used signal decomposition techniques such as Principal Component Analysis (PCA)~\cite{balakrishnan2013detecting} and Independent Component Analysis (ICA)~\cite{poh2010noncontact,monkaresi2014machine} to recover physiological signals under low signal-to-noise ratio conditions. However, these traditional methods were limited by external factors such as head movements and lighting changes, which interfered with their application.

With the rise of deep learning techniques, rPPG measurement tasks have seen significant improvements. Convolutional Neural Networks (CNNs) have been widely applied to skin segmentation and rPPG feature extraction. Early works used 3D CNNs or 2D CNNs to capture spatio-temporal information~\cite{poh2010noncontact,yu2019remote,wang2019vision,chen2018deepphys}, enabling the reconstruction of rPPG signals. In recent years, transformers have been introduced to rPPG tasks~\cite{yu2022physformer,zou2024rhythmformer,yu2023physformer++}, enhancing quasi-periodic rPPG features and global spatio-temporal perception, further improving accuracy. As subtle physiological movements are often affected by external factors, new methods have introduced techniques such as inverse attention mechanisms or temporal shift modules to effectively suppress interference caused by head movements~\cite{nowara2021benefit}.

\vspace{-0.8em}
\subsection{RF radar-Based Methods}
Radar-based remote photoplethysmography (rPPG) technology has evolved significantly since its inception in the 1970s for respiratory-rate detection \cite{lin1975noninvasive}. Over time, radar has been increasingly applied to monitor vital signs like heart rate, respiratory rate, and blood pressure. Various radar systems, including FMCW, UWB Impulse, and Continuous Wave Doppler radars, are used to detect minute chest displacements caused by physiological movements.

Early research showed that radar-based and camera-based methods for heart-rate estimation performed similarly under ideal conditions, but radar systems are more susceptible to noise, often requiring subjects to remain still. Recent advancements have integrated deep learning techniques into FMCW radar to enhance the detection of plethysmograph signals, improving the accuracy of heart-rate estimation~\cite{ha2020contactless}.

In RF-based remote physiology, radar’s ability to capture doppler information and its superior depth resolvability allows it to track subtle oscillations in the chest, providing a precise measurement of vital signs. Initially, most RF techniques relied on signal decomposition methods such as frequency analysis and wavelet decomposition. However, recent studies have leveraged deep learning to further improve the interpretation of radar signals. For example, \cite{ha2020contactless} proposed an encoder-decoder model for reconstructing vital signals from raw RF data, while \cite{zheng2021morefi} employed variational inference.

\vspace{-0.8em}
\subsection{Multi-modal Fusion Methods}
Multi-modal fusion in remote photoplethysmography (rPPG) enhances vital sign estimation performance by combining multiple data modalities. This process involves integrating different modalities to achieve better results than using a single modality. In deep learning, fusion can occur in a middle latent space, where features from different modalities are combined, or at a later decision stage, where predictions from each modality are aggregated. Nevertheless, previous studies have attempted to fuse modalities like RGB with Mid-Infrared (thermal) and Near-Infrared (NIR) to improve rPPG performance~\cite{negishi2020contactless, matsumura2020rgb,park2022self}. Previous studies have also combined RGB and RF to better estimate human vital signs~\cite{vilesov2022blending}. In the field of depression detection, there have been studies that use Bi-SSM as the core framework, combining audio and video information for depression detection.

Additionally, recent research has focused on combining RGB and radar frequency (RF) signals to enhance robustness in challenging conditions, such as low light or adverse weather. This is particularly important for outdoor applications like autonomous driving, where spatial fusion of RGB and RF signals helps with object detection.

\vspace{-0.8em}
\subsection{Mamba}
Mamba \cite{gu2022efficiently} was initially introduced in the field of natural language processing to efficiently handle long sequence data. As research progressed, many variations of mamba have emerged~\cite{fu2023hungry,mehta2023long,smith2023simplified,xie2024fusionmamba,luo2024physmamba}, and Mamba was extended to the computer vision domain, notably through the incorporation of Bidirectional State Space Models (BSSM), resulting in Vision Mamba (Vim)~\cite{zhu2024vision}. Vim processes image sequences bidirectionally and integrates positional embeddings, effectively capturing global visual context and enhancing the efficiency and performance of high-resolution image processing. Compared to traditional transformer models, Vim achieves a 2.8-fold increase in inference speed and reduces GPU memory usage when processing high-resolution images. This advancement signifies a substantial breakthrough for the Mamba architecture in visual tasks.

\begin{figure}
\centering
\includegraphics[width=1\linewidth, height=4cm]{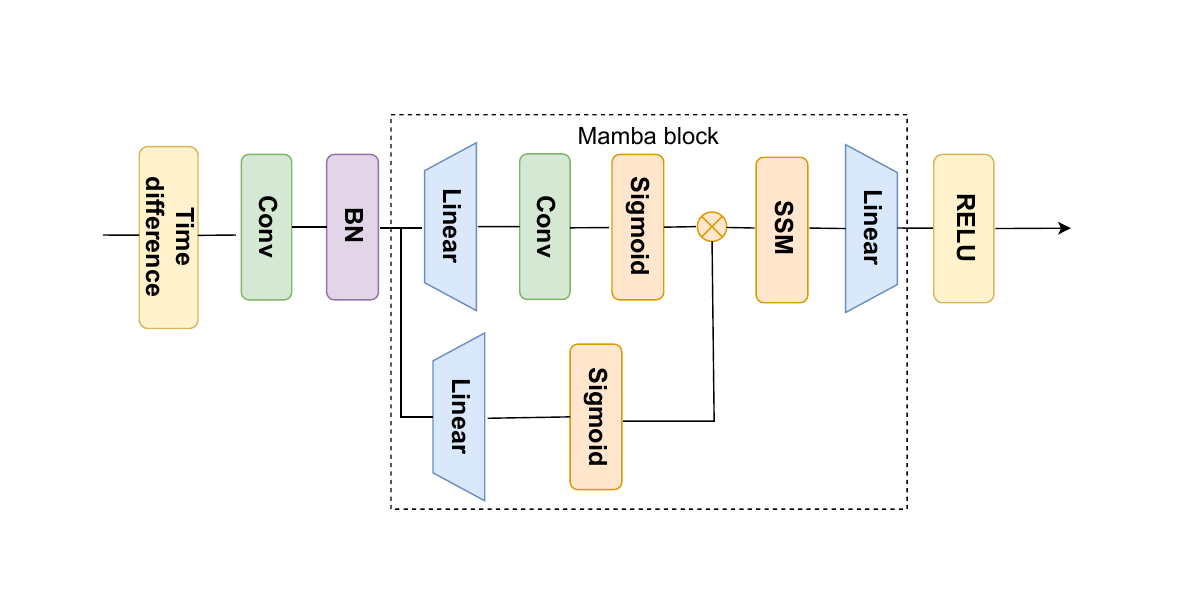}
\vspace{-1.5em}
\caption{Time difference Mamba Module (TDMM) for extracting RF dynamic timing features and global features. }
\label{fig:domain adaptation}
\vspace{-0.2em}
\end{figure}

\vspace{-0.5em}
\section{Methodology}
\label{sec:method}
\subsection{Preliminaries}
\subsubsection{ State Space Representation in the Continuous Domain}

Mamba and S4, both state space model (SSM)-based approaches in deep learning, originate from the concept of linear time-invariant (LTI) systems in classical control theory. An LTI system maps a one-dimensional continuous input sequence $\displaystyle x(t) \in \mathbb{R}$ through an intermediate hidden state $\displaystyle h(t) \in \mathbb{R}^N$ to an output $\displaystyle y(t) \in \mathbb{R}$.
Mathematically, such a system is described by the following linear ordinary differential equation (ODE):
\begin{equation}
  \begin{cases}
h'(t) = A\,h(t) + B\,x(t), \\
y(t) = C\,h(t) + D\,x(t),
\end{cases}
\label{eq1}
\end{equation}
where 
\(\displaystyle A \in \mathbb{R}^{N\times N}\)
is the state transition matrix, 
\(\displaystyle B \in \mathbb{R}^{N\times 1}\) 
and 
\(\displaystyle C \in \mathbb{R}^{1\times N}\)
serve as projection matrices, and 
\(\displaystyle D \in \mathbb{R}\)
denotes a skip connection. While this continuous-time formulation benefits from a solid foundation in control theory, applying it directly to discrete sequence data (e.g., text or speech) in deep learning requires discretization to align with modern hardware computation.

\subsubsection{ Zero-Order Hold (ZOH) Discretization}

To bridge continuous ODEs and discrete sequence modeling, one typically introduces a \emph{time scale parameter} 
\(\displaystyle \Delta\)
and applies the \emph{Zero-Order Hold (ZOH)} principle to discretize the continuous parameters 
\(\displaystyle A\)
and 
\(\displaystyle B\)
into their discrete counterparts 
\(\displaystyle \overline{A}\) 
and 
\(\displaystyle \overline{B}\).
The essential transformation is captured as follows:
\begin{equation}
    \overline{A} = \exp(\Delta A),
\quad
\overline{B} = (\Delta A)^{-1}\Bigl[\exp(\Delta A) - I\Bigr]\,\Delta B,
\end{equation}
where \(\displaystyle \exp(\cdot)\) is the matrix exponential and 
\(\displaystyle I\)
denotes the identity matrix of compatible dimensions. Once discretization is complete, Eq.~\eqref{eq1} can be rewritten on discrete time steps (or sequence indices) as
\begin{equation}
\begin{cases}
h_k = \overline{A}\,h_{k-1} + \overline{B}\,x_k, \\
y_k = C\,h_k + D\,x_k,
\end{cases}
\label{eq3}
\end{equation}
with $\displaystyle k$ indexing each discrete time step or sequence element. This step effectively maps the continuous state dynamics into a discrete iterative framework amenable to deep learning.

\subsubsection{ From RNN Form to Convolutional Form}

While Eq.~\eqref{eq3} superficially resembles a class of RNN hidden-state updates, it can be further recast into a one-dimensional convolutional form. Specifically, by unrolling $\displaystyle h_k$ over preceding time steps and collecting terms of $\displaystyle \{\overline{A}^{k-1}\,\overline{B}\}$, one can define a \emph{structured convolutional kernel} $\displaystyle \overline{K}$, yielding a single convolution operation over the sequence $\displaystyle \{x_1, x_2, \dots, x_L\}$:
\begin{equation}
\overline{K} = 
\bigl(
C\,\overline{B}, \;
C\,\overline{A}\,\overline{B}, \;\dots,\;
C\,\overline{A}^{\,L-1}\,\overline{B}
\bigr),
\quad 
y = x \,\ast\, \overline{K},
\end{equation}
where $\displaystyle \ast$ denotes one-dimensional convolution and 
$\displaystyle L$ is the sequence length. This formulation confers significant parallelization advantages, leveraging large-scale GPU/TPU parallelism to greatly enhance efficiency and scalability for long-sequence modeling.

\begin{figure}
\centering
\includegraphics[width=1\linewidth, height=4cm]{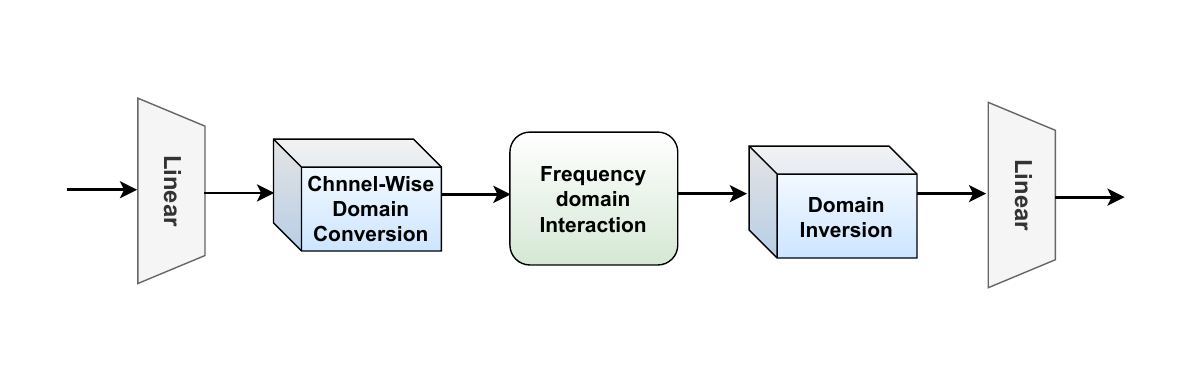}
\vspace{-1.5em}
\caption{Channel-wise Fast Fourier Transform
(CFFT) is used to extract frequency domain features of RGB and RF modalities.}
\label{fig:domain adaptation2}
\vspace{-0.2em}
\end{figure}

\subsection{Overview}

The proposed CardiacMamba model incorporates three fundamental components for the modality fusion process: Dual-level Feature Extraction and Alignment, Bidirectional Feature Interaction, and Bidirectional Feature Fusion. As shown in Fig.~\ref{fig:network}, the RGB modality input is denoted as \( I_{C} \in \mathbb{R}^{3 \times T_{\text{1}} \times H \times W} \), while the RF modality input is represented as \( I_{f} \in \mathbb{R}^{C \times T_{\text{2}}} \).

The initial phase, Dual-level Feature Extraction and Alignment, encompasses both low-level and high-level feature extraction processes. In the low-level stage, we employ Temporal Difference
Mamba Module (TDMM) and Bifurcated Diff-Conv Fusion (BDCF) to capture primary features from both the RGB and RF modalities. The RGB modality output, \( H_{c}^{n_{1}} \), is then processed by Spatial-Channel Fusion Module (SCFM), which extracts rPPG-related features and consolidates spatial information into token channels. For the RF modality output, \( H_{f}^{n_{1}} \), we utilize two  RF Alignment Module (RFAM) modules to extract deeper features and align the RF modality with the RGB modality in the temporal domain.
\begin{equation}
H_{c}^{n_{2}} = SCFM(BDCF(I_{C}))
\end{equation}
\vspace{-1.0em}

\begin{equation}
H_{f}^{n_{2}} = RFAM(RFAM(TDMM(I_{C})))
\end{equation}

In the subsequent Bidirectional Feature Interaction stage, the outputs \( H_{c}^{n_{2}} \) and \( H_{f}^{n_{2}} \) from the RGB and RF modalities are fed into the Vision Mamba (Vim) module, which aligns both modalities into a shared representation space, facilitating the interaction of information across modalities and enabling the extraction of global features.

The aligned features are then fused by adding \( H_{c}^{n_{2}} \) to \( Vim(H_{c}^{n_{2}}) \) for the RGB modality, and similarly adding \( H_{f}^{n_{2}} \) to \( Vim(H_{f}^{n_{2}}) \) for the RF modality. These fused features are subsequently transformed into a new feature space through a linear transformation, resulting in \( H_{c}^{n_{4}} \) and \( H_{f}^{n_{4}} \).

This fusion process enhances the exchange of information between modalities, thereby producing a more comprehensive feature representation that improves the accuracy and robustness of the final predictions.

\begin{equation}
H_{c}^{n_{4}} = \text{Linear}(H_{c}^{n_{2}} + Vim(H_{c}^{n_{2}}))
\end{equation}

\begin{equation}
H_{f}^{n_{4}} = \text{Linear}(H_{f}^{n_{2}} + Vim(H_{f}^{n_{2}}))
\end{equation}

Subsequently, the fused outputs from the Vim module are passed through a channel-wise Fast Fourier Transform (CFFT) network, which further facilitates the interaction of information across multiple channels. Finally, the aggregated features are forwarded to the predictor to estimate the BVP signal.

\begin{equation}
H_{c}^{n_{5}}, H_{f}^{n_{5}} = \text{CFFT}(H_{c}^{n_{4}}), \text{CFFT}(H_{f}^{n_{4}})
\end{equation}

\subsection{Dual-level Feature Extraction and Alignment}

\subsubsection{Low-level Feature Extraction.}

\textbf{Temporal Difference Mamba Module (TDMM)}: To better capture the dynamic changes in the time series of the radio frequency (RF) modality, we propose a novel module called the Temporal Difference
Mamba Module (TDMM). As shown in Fig.~\ref{fig:domain adaptation} and Algorithm 1, This module is designed to effectively handle time difference information and extract more discriminative features from the temporal data, thereby significantly enhancing the temporal information in the RF signal and improving the model’s performance in complex dynamic scenarios.

 Firstly, we calculate the differences between consecutive frames in the input RF data to help the network capture the dynamic changes of the signal. Specifically, for each frame, we apply a temporal shift operation to obtain $X_{t-2}, X_{t-1}, X_{t}, X_{t+1}, X_{t+2}$ (i.e., five adjacent frames). By calculating the differences between each frame and its preceding and succeeding frames, we obtain frame differences $D_{-2}, D_{-1}, D_{1}, D_{2}$, which reflect the variation characteristics of the RF signal at different time points.
    
Next, we combine the differential RF signal with the original signal, then we apply convolution operations and batch normalization to these time-differenced maps. Given that RF signals typically contain periodic dynamic features and to avoid introducing excessive redundancy in the temporal dimension, we use 7×1 convolution kernels to efficiently extract these features. We then introduce the Mamba block to further extract global features, and finally apply the ReLU activation function to enhance the non-linear expression.

\textbf{Bifurcated Diff-Conv Fusion (BDCF)}: The diff-fusion module first integrates frame differences into the raw frames, enhancing the perception of BVP wave variations. This effectively improves rPPG features with minimal computational cost. Since rPPG requires both high-frequency information across frames and low-frequency information within frames, large convolutional kernels are used to capture the low-frequency information within frames, fully incorporating spatial information into the channels.

\begin{algorithm}[t]
	\renewcommand{\algorithmicrequire}{\textbf{Input:}}
	\renewcommand{\algorithmicensure}{\textbf{Output:}}
	\caption{Temporal Difference Mamba Module (TDMM).     \textbf{Notation:} `B' (Batch Size), `C'  (Number of Channels), `T' (Number Of Frames), `D' (Frame Difference)}
	\label{tab:comparison} 
	\begin{algorithmic}[1]
    
     \REQUIRE $I_{f}:(B,C_{i},T_{i})$
    \ENSURE $X_{3}: (B,C_{i},T_{i})$
		\STATE   $X_{t-2}, X_{t-1}, X_t, X_{t+1} \leftarrow \text{Temporal\ Slice}(I_{f})$.
        \STATE   \small$D_{-2}, D_{-1}, D_{1}, D_{2} \leftarrow $ Compute difference between adjacent slices.
        \STATE   $X_{diff} \leftarrow Concatenate(D_{-2}, D_{-1}, D_{1}, D_{2})$.
            \FOR{$k=1$ to $N$}
            \STATE $X:(B,C_{i},T_{i})  \leftarrow Conv(X_{diff})$. 
            \STATE $X':(B,C_{i},T_{i}) \leftarrow Batchnorm(X)$. 
            \\ \texttt{\# Mamba Block}.
            \STATE $X'':(B,C_{i},T_{i}) \leftarrow Linear(X')$.
            \STATE $X''\  \leftarrow Conv(X'')$.
            \STATE $\overline{X}:(B,C_{i},T_{i}) \leftarrow Linear(X')$.
            \STATE $\overline{X}:(B,C_{i},T_{i})\  \leftarrow Sigmoid(\overline{X})$.
            \STATE $X''' :(B,C_{i},T_{i}) \leftarrow \overline{X} \odot X''$.
            \STATE $\overline{X1}:(B,C_{i},T_{i})  \leftarrow SSM(X''')$.
            \STATE $X2:(B,C_{i},T_{i}) \leftarrow Linear(X1)$.
            \STATE $X3:(B,C_{i},T_{i}) \leftarrow RElu(X2)$.
            \ENDFOR
	\end{algorithmic}  
\end{algorithm}

The process is as follows: For an input video \( X \in \mathbb{R}^{3 \times T \times H \times W} \), temporal shifting is applied to obtain \( X_{t-2}, X_{t-1}, X_t, X_{t+1} \), and \( X_{t+2} \). The differences between consecutive frames are then computed, resulting in \( D_{-2}, D_{-1}, D_{1} \), and \( D_2 \). These frame differences are combined with the raw frames and passed through \( \text{Stem1} \) for feature extraction. \( \text{Stem1} \) consists of a \( (7 \times 7) \) convolution layer, followed by batch normalization (BN), ReLU, and max pooling. The input dimension is 3 when using single-frame features, and 12 when concatenating frame differences.

 Next, the features of the difference video and the raw video are merged and further enhanced through \( \text{Stem2} \), which also includes a \( (7 \times 7) \) convolution layer, BN, and ReLU.The fusion formula is:
\[
X_{fu} = \alpha \cdot \text{Stem2}(X_{origin}) + \beta \cdot \text{Stem2}(\alpha \cdot X_{origin} + \beta \cdot X_{diff})
\]

The fusion coefficients \( \alpha \) and \( \beta \) are both set to 0.5. The output of the BDCF is a fused and enhanced feature representation.

\subsubsection{High-level Feature Extraction}

\textbf{ Spatial-Channel Fusion Module (SCFM)}: This module aims to effectively capture high-level features from RGB video signals.
\begin{table*}[h]
\centering
\renewcommand{\arraystretch}{1.3}  
\small  
\setlength{\tabcolsep}{14pt}  
\caption{Comparison of different methods on the EquiPleth dataset. The best results are marked in \textbf{bold}.}
\begin{tabular}{l|cccc}
\hline
Method & Input & MAE & RMSE & $\rho$ \\ \hline
DeepPhys~\cite{chen2018deepphys} & RGB & 5.54 & 18.51 & 0.66 \\ 
PhysNet~\cite{yu2019remote} & RGB & 8.06 & 19.71 & 0.61 \\ 
MTTS-CAN ~\cite{liu2020multi} & RGB & 3.69 & 13.8 & 0.82 \\ 
PhysFormer~\cite{yu2022physformer} & RGB & 12.92 & 24.36 & 0.47 \\ 
EfficientPhys~\cite{liu2023efficientphys} & RGB & 5.47 & 17.04 & 0.71 \\ 
RhythmMamba~\cite{zou2024rhythmmamba} & RGB & 2.87 & 9.58 & 0.92 \\ 
\textbf{Ours (RGB-Only)} & RGB & 1.2 & 4.23 & 0.95 \\ \hline
Tu et al.~\cite{tu2016respiration} & RF & 5.5 & 11.68 & 0.64 \\ 
Mercuri et al.~\cite{mercuri2019vital} & RF & 4.73 & 9.6 & 0.7 \\  
FFT-based \cite{alizadeh2019remote} & RF & 13.51 & 21.07 & 0.24 \\ 
\textbf{Ours (RF-Only)} & RF & 5.2 & 7.4 & 0.8 \\ \hline
Vilesov et al. \cite{vilesov2022blending} & RGB+RF & 1.12 & 3.42 & 0.95  \\ 
\textbf{Ours (Full model)} & RGB+RF & \textbf{0.96} & \textbf{3.06} & \textbf{0.97} \\ 
\hline
\end{tabular}
\label{tab:comparison}
\end{table*}
First, the input features undergo an initial nonlinear transformation through the Sigmoid activation function. The use of the Sigmoid function compresses the input signal to the range of [0, 1], enhancing model stability and effectively suppressing outliers.  The activated features are then passed into the Attention Mask module, which uses a self-attention mechanism to weight the input signal and capture the relationships between different spatial locations. Specifically, the Attention Mask is computed as follows:

\begin{equation}
Mask = \frac{(H/8)(W/8) \cdot \sigma(\text{Stem}(X_{\text{fu}}))}{2 \| \sigma(\text{Stem}(X_{\text{fu}})) \|_1},
\end{equation}
where $\sigma$ represents the Sigmoid activation function, and $\text{Stem}$ is a relatively large (5×5) convolution kernel that effectively integrates spatial information into the channel. In this way, the Attention Mask highlights the strong signal regions in the skin areas of the image, which is crucial for tasks like heart rate estimation. Compared to traditional Softmax mechanisms, L1 normalization ($ | \cdot |_1 $) is smoother and produces a sparser mask, allowing the Attention Mask to more precisely focus on the key signal regions, while requiring fewer computational resources.




The generated Attention Mask is then applied to the fused features $X_{\text{fusion}}$, resulting in an enhanced feature representation $X_{\text{attn}} \in \mathbb{R}^{C \times T / 2 \times H / 8 \times W / 8}$. This enhancement process effectively reduces unnecessary noise and strengthens the representation of the signal in the spatial domain through L1 normalization.

Next, the features processed by the self-attention mechanism undergo global average pooling to further integrate the features. Global average pooling reduces the feature dimensionality while effectively incorporating spatial information into the channels, making the representation for each channel more compact and representative. The pooled output features $X_{\text{stem}} \in \mathbb{R}^{T / 2 \times C}$ provide highly condensed information for the subsequent convolution operations.

Finally, the pooled features are processed through convolution and batch normalization layers. The convolution operation further extracts useful information from the feature space, enhancing the model's expressiveness, while batch normalization standardizes the feature distribution across channels, accelerating convergence and improving training stability.

    \textbf{RF Alignment Module (RFAM)}: RF Alignment Module (RFAM) module is designed to process RF modality data by enhancing its feature representation and ensuring temporal alignment with other modalities, such as RGB. The module consists of several critical operations: convolution, batch normalization, channel attention, downsampling, and ReLU activation. Initially, the RF data is passed through convolutional layers to extract spatial features, followed by batch normalization to standardize the feature distributions across the mini-batch. 

Next, the processed features undergo a channel attention mechanism that adaptively highlights the most informative channels. The attention mechanism uses both average pooling and max pooling to capture global spatial information, which is then passed through a fully connected network comprising two convolutional layers. The attention weights are calculated using a Sigmoid activation, and these weights are applied to the input feature map, selectively amplifying the relevant channels while suppressing less useful ones. This step allows the model to focus on the most important features of the RF data, which is particularly useful for handling the dynamic nature of RF signals.

Finally, the RF data undergoes downsampling to align its temporal dimension with that of the RGB data. This temporal alignment ensures that both modalities are synchronized and can be effectively fused in subsequent stages of the model. After downsampling, the data is passed through a ReLU activation function, which introduces non-linearity and helps capture complex patterns in the features.

\subsection{Bidirectional Feature Interaction}
Inspired by \cite{ye2024depmamba,zhu2024vision}, we proposed Bidirectional Feature Interaction enhances multimodal representation by collaboratively modeling RF and video features. With the Vision Mamba encoder \cite{ye2024depmamba}, it captures temporal dependencies between the modalities and facilitates their interaction via a shared state transition matrix, without adding parameters. This approach allows both modalities to learn cross-modal information while preserving their unique characteristics, improving the model's expressiveness and accuracy.

The Vision Mamba Encoder first applies linear projection to transform each image patch into a sequence of tokens, augmented with positional embeddings to maintain spatial relationships. It then models the patch sequence bidirectionally through a Bidirectional State Space Model (Bi-SSM), capturing comprehensive spatiotemporal dependencies. This bidirectional modeling integrates both global context and local details
\begin{table}
    \centering

    \renewcommand\arraystretch{1} 
    \setlength{\tabcolsep}{2.5mm} 
    \caption{Comparison of the performance of various methods on the RGB and RF fusion task, with the differences in performance ($\Delta$) between light and dark skin tones in terms of MAE, RMSE, and correlation coefficient ($\rho$). }
    \label{tab:comparisonb} 
    \resizebox{0.48\textwidth}{!} { 
        \begin{tabular}{@{}lllll@{}}
            \toprule
            Method & Input & $\Delta$MAE & $\Delta$RMSE & $\Delta$$\rho$ \\ 
            \midrule
            ICA \cite{poh2010advancements} & RGB & 4.42 & 3.15 & -0.36 \\ 
            CHROM \cite{dehaan2013robust} & RGB & 4.97 & 4.17 & -0.38 \\ 
            BCG \cite{balakrishnan2013detecting}& RGB & 0.99 & \textbf{1.25} & \textbf{0.05} \\ 
            PhysNet \cite{yu2019remote} & RGB & 2.22 & 4.05 & -0.25 \\ 
            FFT-based RF \cite{alizadeh2019remote} & RF & 1.32 & 2.06 & 0.32 \\
           Vilesov et al. \cite{vilesov2022blending} & RGB\&RF & 0.67 & 1.44 & -0.10 \\ 
            \textbf{CardiacMamba (Ours)} & RGB\&RF & \textbf{0.26} & 1.28 & \textbf{0.05} \\ 
            \bottomrule
        \end{tabular}
    }
\end{table}
effectively. Convolutional operations extract local features, helping the model identify fine-grained patterns, while global convolution layers capture long-range dependencies. This multi-level structure generates rich visual representations for downstream tasks such as classification and object detection.

In multimodal tasks, the Vision Mamba Encoder excels with both RGB video and RF radar modalities. For RGB video, bidirectional modeling captures temporal dependencies, aiding in dynamic information extraction like facial expressions and pose changes. For RF radar, the low resolution and noise typically hinder detail extraction. However, by combining RF signals with RGB video signals, bidirectional modeling and shared state matrices enhance the RF signal’s temporal dynamics, improving multimodal fusion and model performance.

Unlike traditional multimodal learning, which uses separate state transition matrices for each modality, Bidirectional Feature Interaction introduces a shared state transition matrix. This unified approach ensures that both RF and video modalities evolve under the same dynamic rules, fostering closer collaboration. The shared matrices A and B allow both modalities to update simultaneously, enhancing fusion and complementarity:
\begin{equation}
h_c^{(t+1)} = A h_c^{(t)} + B x_c^{(t)}
\end{equation}

\begin{equation}
h_f^{(t+1)} = A h_f^{(t)} + B x_f^{(t)}
\end{equation}

This shared matrix framework simplifies the model and improves cross-modal integration, enabling more efficient learning and better performance on multimodal tasks.

The introduction of a shared state transition matrix not only simplifies the model structure but also significantly improves the efficiency of cross-modal information integration. By sharing matrices \( A \) and \( B \), Bidirectional Feature Interaction can capture the interdependencies between RF signals and video images under the same temporal dynamics. The shared matrices eliminate redundancy between modalities, facilitate information exchange and collaborative learning between modalities, and thus effectively improve the performance of multimodal learning tasks.

\subsection{Bidirectional Feature Fusion}
In this subsection, the features of the RGB and RF modalities are first fed into their respective Channel-wise Fast Fourier Transforms (CFFT) to extract higher-order frequency-domain features. The features from both modalities are then combined through summation, and the fused features are passed into the predictor for decoding to obtain the rPPG signal.

As shown in Fig.~\ref{fig:domain adaptation2}, the CFFT module first applies a fast Fourier transform to map the channel-domain signals into the frequency domain, highlighting the frequency information of different modalities across channels. It then uses learnable interactions between the real and imaginary parts in the frequency domain to enhance or suppress specific frequency components. Finally, by applying the inverse Fourier transform, the processed results are restored back to the channel domain. This process preserves the original feature representations while effectively fusing RGB and RF signals in the frequency domain and strengthening the model’s ability to detect heart rate periodicity.

By performing a Fast Fourier transform along the channel dimension rather than directly on the time dimension or other dimensions, we can effectively extract the unique frequency characteristics of each modality, since the RGB and RF signals carry distinct frequency-domain information across different channels. This channel-wise approach allows the model to capture those distinct frequency components of both RGB and RF signals and to uncover their cross-modal frequency relationships, thereby improving heart rate estimation accuracy. In addition, it can enhance or suppress specific frequency components to remove noise more effectively and highlight those related to heart rate.
\begin{table}
\centering
 
\renewcommand{\arraystretch}{1.1}  
\small  
\setlength{\tabcolsep}{5pt}  
\caption{Testing under missing-modality conditions.}
\label{tab:comparison3}
\begin{tabular}{@{}llllll@{}}
\hline

            Method & Train & Test & MAE & RMSE  \\ 
            \midrule
            Base & RGB\&RF & RGB & 21.7 & 25.7  \\ 
            Base & RGB\&RF & RF & 21.4 & 24.3  \\ 
            Base & RGB\&RF & RGB\&RF & 20.3 & 24.8  \\ \hline
            Vilesov et al.~\cite{vilesov2022blending} & RGB\&RF & RGB & 6.82 & 13.32  \\ 
            Vilesov et al.~\cite{vilesov2022blending} & RGB\&RF & RF & \textbf{7.25} & \textbf{9.62}  \\ 
            Vilesov et al.~\cite{vilesov2022blending} & RGB\&RF & RGB\&RF & 1.12 & 3.42  \\ \hline 
            \textbf{CardiacMamba (Ours)} & RGB\&RF & RGB & \textbf{1.2} & \textbf{3.41}  \\ 
            \textbf{CardiacMamba (Ours)} & RGB\&RF & RF & 11 & 13  \\ 
            \textbf{CardiacMamba (Ours)} & RGB\&RF & RGB\&RF & \textbf{0.96} & \textbf{3.06}  \\ 
            \bottomrule
        \end{tabular}
\vspace{-1.5em} 
\end{table}

Fast Fourier Transform (FFT) along the Channel Dimension: We apply the FFT to the features along the channel dimension, transforming the signal from the time domain into the frequency domain:

\begin{equation}
      \small
\begin{aligned}
H(f_c)
&= \int_{-\infty}^{\infty} H(c)\,e^{-j2\pi f_c\,c}\,\mathrm{d}c \\
&= \int_{-\infty}^{\infty} H(c)\cos\bigl(2\pi f_c\,c\bigr)\,\mathrm{d}c
\;+\; j\int_{-\infty}^{\infty} H(c)\sin\bigl(2\pi f_c\,c\bigr)\,\mathrm{d}c \\
&= H(f_c)_\mathrm{re} \;+\; j\,H(f_c)_\mathrm{im},
\end{aligned}
\end{equation}
where \( c \) is the channel dimension (continuous variable in this integral form),
\( f_c \) is the channel-frequency variable,
\( H(c) \) denotes the original signal in the channel domain,
\( H(f_c)_\mathrm{re}, H(f_c)_\mathrm{im} \) are the real and imaginary parts of \( H(f_c) \),
and \( j \) is the imaginary unit with \( j^2 = -1 \).

Frequency-domain Interaction: After transforming the features into the frequency domain, we perform interaction operations using learnable parameters. This enables the model to focus on the most important frequency components and suppress irrelevant ones. The interaction between the real and imaginary parts of the frequency-domain features is modeled as:

\begin{equation}
H_{\text{real}} = \text{ReLU}\left( \sum_{b=0}^{B} \left( X_{\text{fre}}^{\text{real}} r - X_{\text{fre}}^{\text{imag}} i + r_b \right) \right)\end{equation}

\begin{equation}H_{\text{imag}} = \text{ReLU}\left( \sum_{b=0}^{B} \left( X_{\text{fre}}^{\text{imag}} r + X_{\text{fre}}^{\text{real}} i + i_b \right) \right)\end{equation}

Inverse Fourier Transform (IFFT): After the frequency-domain interaction, the inverse Fourier Transform (IFFT) is applied to convert the features back to the time domain:

\[
\begin{aligned}
H(c)
&= \int_{-\infty}^{\infty} H(f_c)\,e^{\,j2\pi f_c\,c}\,\mathrm{d}f_c \\
&= \int_{-\infty}^{\infty} 
\bigl(H(f_c)_\mathrm{re} \;+\; j\,H(f_c)_\mathrm{im}\bigr)\,
e^{\,j2\pi f_c\,c}\,\mathrm{d}f_c.
\end{aligned}
\]

\subsection{Loss}
\text{Our loss includes a Negative Pearson Loss $L_{neg}$} \text{and a Signal-to-Noise Ratio (SNR) loss $L_{SNR}$}
\text{The Negative Pearson Loss is defined as follows:}
\begin{equation}
 \mathcal{L}_{n}(y, \hat{y}) = 1 - \frac{1}{\sqrt{a_1 \times a_2}} \left( \sum_{i=1}^{N} y_i \hat{y}_i - \sum_{i=1}^{N} y_i \hat{y}_i \right) 
\end{equation}

\begin{equation}
\label{eq15}
 a_1 = \left( \sum_{i=1}^{N} y_i^2 - \left( \sum_{i=1}^{N} y_i \right)^2 \right)
\end{equation}
\begin{equation}
 a_2 = \left( \sum_{i=1}^{N} \hat{y}_i^2 - \left( \sum_{i=1}^{N} \hat{y}_i \right)^2 \right)
\end{equation}

The SNR loss is defined as follows:

\begin{equation}
\mathcal{L}_{SNR}(y, \hat{y}) = \frac{ \int_{f_0 - w}^{f_0 + w} |\hat{Y}(f)|^2 df }{ \int_{-\infty}^{f_0 - w} |\hat{Y}(f)|^2 df + \int_{f_0 + w}^{\infty} |\hat{Y}(f)|^2 df },
\end{equation}
\begin{equation}
f_0 = \arg \max Y(f),
\end{equation}
where $Y(f)$ and $\hat{Y}(f)$ are the respective Fourier transforms of $y$ and $\hat{y}$ and w is the chosen window size.The overall loss is:
\begin{equation}
    \mathcal{L}(y, \hat{y}) = \mathcal{L}_{neg}(y, \hat{y}) + \lambda \mathcal{L}_{SNR}(y, \hat{y})
\end{equation}

\begin{table}
    \centering
    \small
    \caption{Ablation experiments on different modules, including `Vim' (short for Vision Mamba), `CFFT' (short for Channel-wise Fast Fourier Transform), `RFAM' (RF Alignment Module), and TDMM'`(Temporal Difference Mamba Module). }
    \label{tab:comparison4}
    \renewcommand\arraystretch{1.0} 
    \setlength{\tabcolsep}{1.3mm} 
    {
        \begin{tabular}{@{}llllllll@{}}
            \toprule
            Vim & CFFT & SSM & \makecell[c]{RFAM} & \makecell[c]{TDMM} & MAE & RMSE & $\rho$ \\ 
            \midrule
            $\times$ & \checkmark & \checkmark & \checkmark & \checkmark & 1.7 & 4.9 & 0.91 \\ 
            \checkmark & $\times$ & \checkmark & \checkmark & \checkmark & 18.8 & 20.5 & 0.2 \\ 
            \checkmark & \checkmark & $\times$ & \checkmark & \checkmark & 1.86 & 5.51 & 0.89 \\ 
            \checkmark & \checkmark & \checkmark & $\times$ & \checkmark & 1.85 & 5.31 & 0.9 \\ 
            \checkmark & \checkmark & \checkmark & \checkmark & $\times$ & 3.82 & 8.06 & 0.81 \\
            \checkmark & \checkmark & \checkmark & \checkmark & \checkmark & \textbf{0.96} & \textbf{3.06} & \textbf{0.97}   \\
            \bottomrule
        \end{tabular}
    }
    \vspace{-1.0em} 
\end{table}

\section{Experiment}
\label{sec:experiment}
\subsection{Datasets and Metrics}
\textbf{Datasets.}\quad To conduct evaluations, we performed extensive experiments on the EquiPleth dataset \cite{vilesov2022blending}. The EquiPleth dataset comprises The dataset contains 28 light, 49 medium, and 14 dark skin tone volunteers. The RGB camera, set to default factory settings at 30 fps, and the FMCW radar, set to a starting frequency of 77 GHz, were used to record 6 sessions per volunteer, each lasting 30 seconds and consisting of both RGB video and radar IQ data.

\textbf{Metrics.}\quad This study adopts the evaluation metrics of mean absolute error (MAE), root mean square error (RMSE), and Pearson correlation coefficient ($\rho$).In particular, lower MAE and RMSE values indicate smaller error margins, while $\rho$ values closer to 1.0 reflect reduced error. Both MAE and RMSE are expressed in beats per minute (bpm), and for convenience, these units will be omitted in the following tables and analyses.
\subsection{Experimental Setup}

\label{sec:imp}

The inputs for the RGB branch consisted of video clips that were processed to 128x128 crops using a MTCNN ~\cite{zhang2016joint} to locate facial regions. The inputs for the RF branch consisted of IQ data, which was processed into a range matrix, with data related to regions of interest extracted within a 25 cm window.The proposed Fusion-Vital model was trained using the ADAM optimizer with a batch size of 32, a learning rate of 0.0001, on a 4090 GPU for 30 epochs.

\subsection{Comparison with State-of-the-Art method}
As shown in TABLE~\ref{tab:comparison}, we evaluated the performance of the proposed  model against several state-of-the-art remote physiology models  \cite{chen2018deepphys}~\cite{yu2019remote}~\cite{liu2020multi}~\cite{yu2022physformer}~\cite{liu2023efficientphys} ~\cite{zou2024rhythmmamba}), which depend on a single modality. We also evaluate the performance of our model with multimodal remote physiology model\cite{vilesov2022blending}.

For heart rate estimation, our model outperforms the baseline models by a significant margin. Compared to the best output of the RGB model, our multimodal fusion model reduces MAE by 67\% and RMSE by 68\%. It also outperforms the best RF model. Additionally, when compared to the previously best-performing multimodal fusion model, our approach reduces MAE by 15\% and RMSE by 11\%. While the performance of our single-modality RGB model is comparable to the best output of the RGB model, our single-modality RF model surpasses the performance of the best RF model.

\begin{figure*}[t] 
\centering

\begin{subfigure}[t]{0.23\textwidth} 
    \centering
    \includegraphics[width=\linewidth, height=5cm]{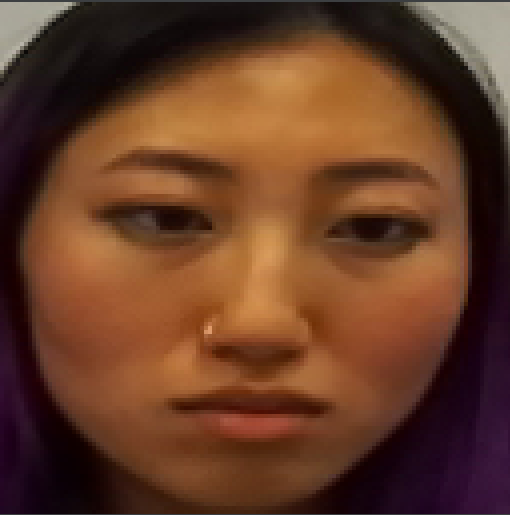} 
    \caption{Human face}
    \label{fig:face_raw1}
\end{subfigure}
\hfill 
\begin{subfigure}[t]{0.23\textwidth}
    \centering
    \includegraphics[width=\linewidth, height=5cm]{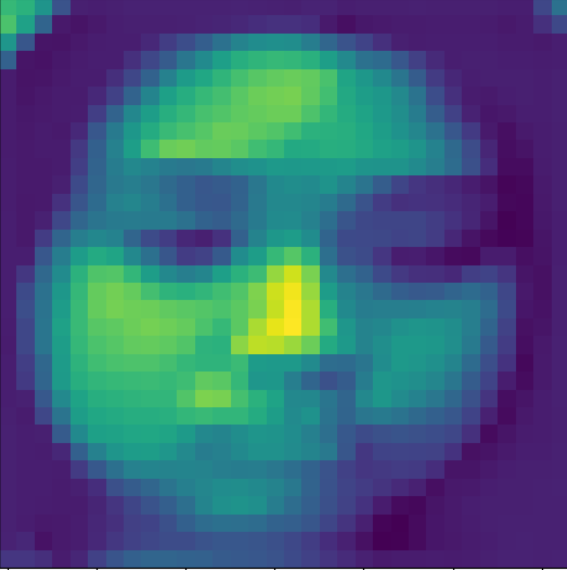}
    \caption{Feature heat map of human face}
    \label{fig:face_af2}
\end{subfigure}
\hfill
\begin{subfigure}[t]{0.23\textwidth}
    \centering
    \includegraphics[width=\linewidth, height=5cm]{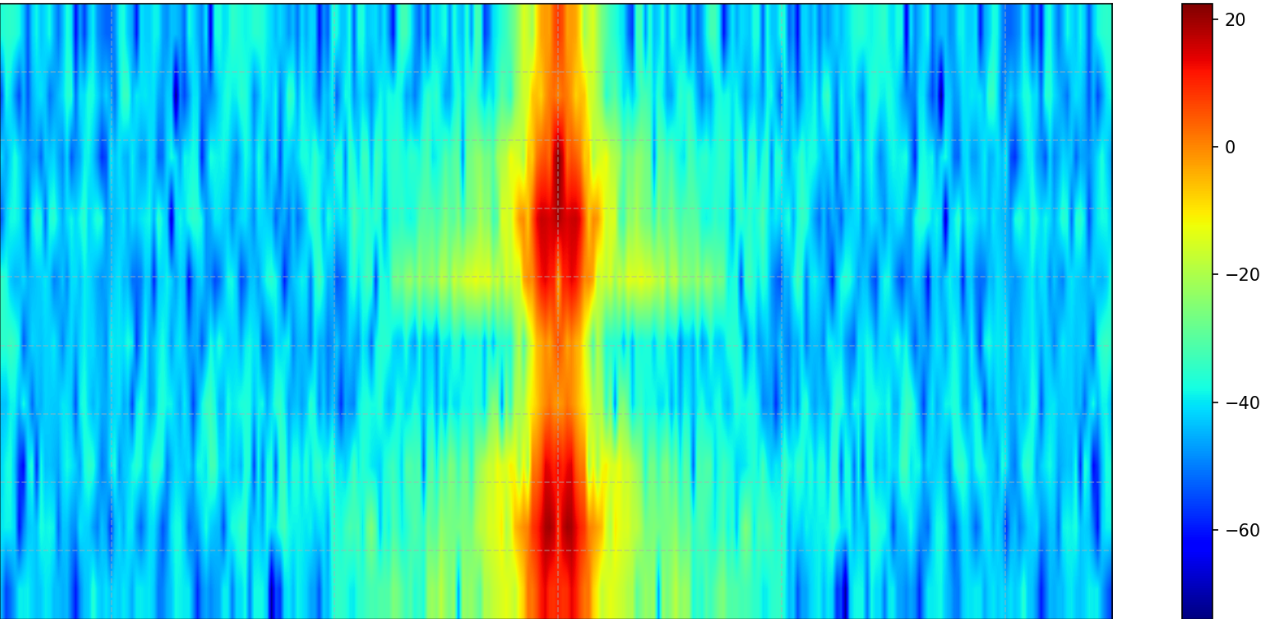}
    \caption{Radar spectrum diagram}
    \label{fig:rf_raw3}
\end{subfigure}
\hfill
\begin{subfigure}[t]{0.23\textwidth}
    \centering
    \includegraphics[width=\linewidth, height=5cm]{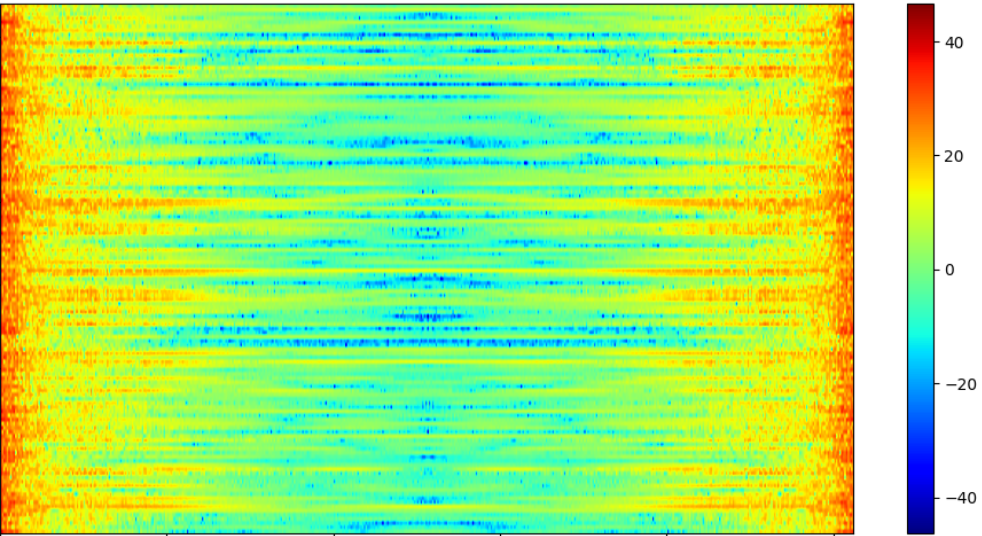}
    \caption{Feature heat map of radar spectrum diagram}
    \label{fig:rf_af4}
\end{subfigure}

\caption{Visual representations of human face and radar spectrum features. (a) The human face image used in the analysis. (b) Feature heat map of the human face, illustrating the key regions of interest. (c) Radar spectrum diagram representing the frequency information of the signal. (d) Feature heat map of the radar spectrum diagram, highlighting the relevant features for analysis.}
\label{fig:full_page_feature1}
\vspace{-0.5em} 
\end{figure*}

\begin{figure}[t]
\centering
\captionsetup{justification=centering}
\begin{subfigure}[t]{0.24\textwidth} 
    \centering
    \includegraphics[width=\linewidth, height=4cm]{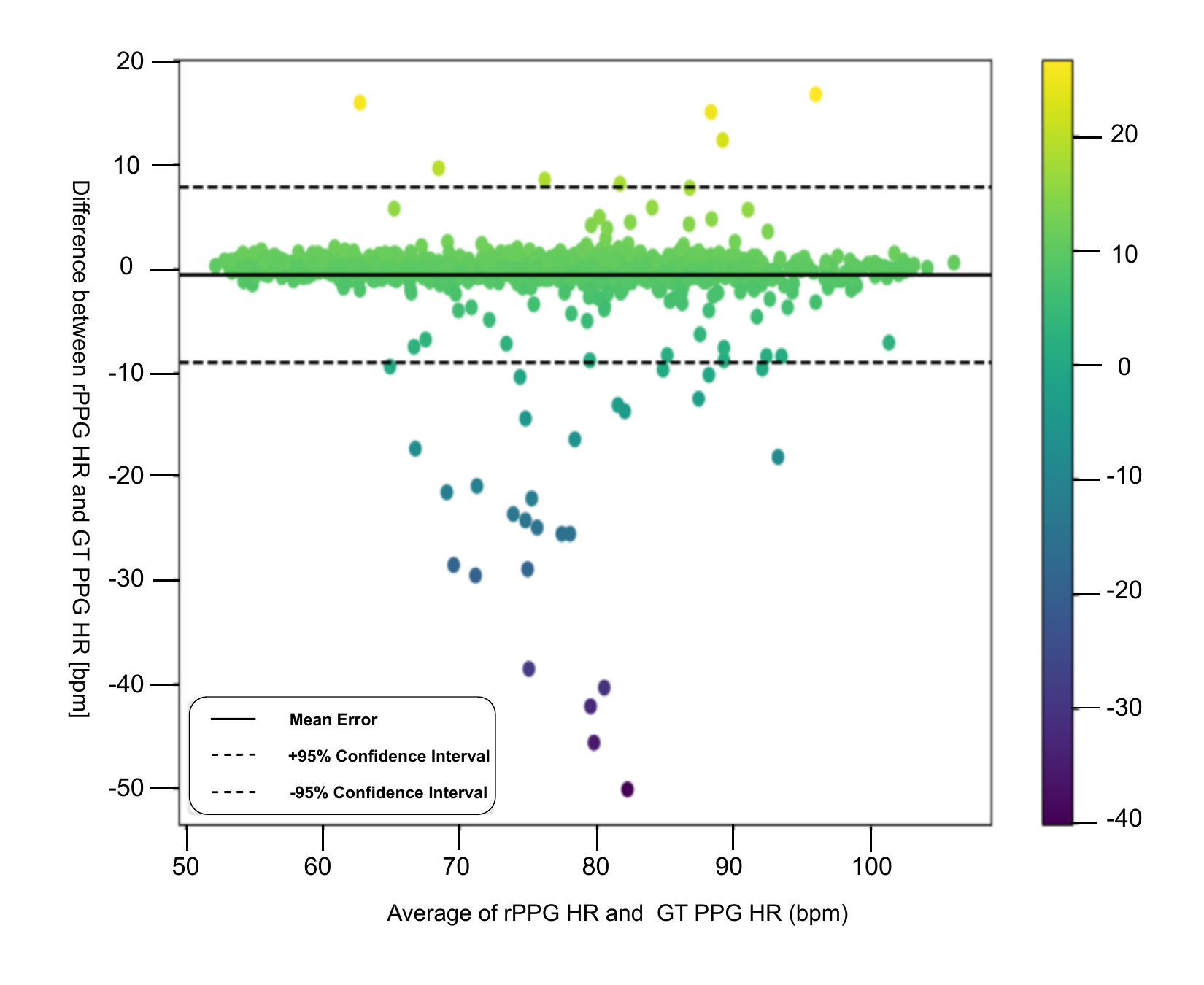}
    \caption{CardiacMamba (Ours)} 
    \label{fig:face_raw2}
\end{subfigure}
\hfill
\begin{subfigure}[t]{0.24\textwidth}
    \centering
    \includegraphics[width=\linewidth, height=4cm]{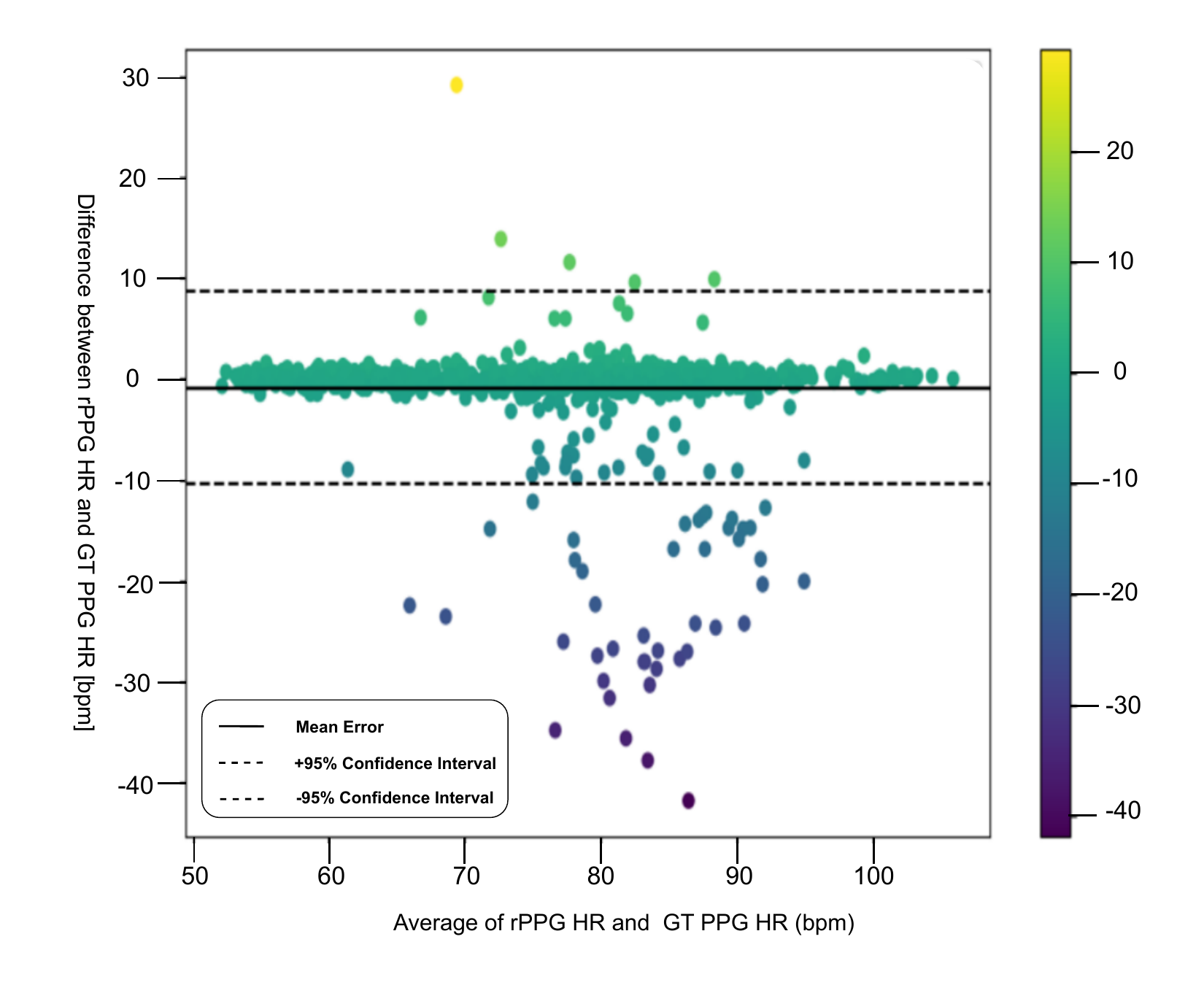}
    \caption{Vilesov et al. \cite{vilesov2022blending}} 
    \label{fig:face_af2}
\end{subfigure}

\caption{Bland-Altman plots comparing the heart rate estimation from different methods. (a) CardiacMamba, showing the agreement between estimated and ground truth (GT) heart rates. (b) Vilesov et al. \cite{vilesov2022blending}, illustrating the discrepancies between estimated and GT heart rates. The color gradient represents the distribution of differences, with confidence intervals indicated by the dashed lines.}
\label{fig:full_page_feature2}
\vspace{-0.5em} 
\end{figure}

\subsection{Measuring Skin Tone Bias and Fairness}

To address skin tone bias, this study systematically evaluates the fairness of different models by quantifying performance differences (Difference Value) between light and dark skin samples. As shown in TABLE~\ref{tab:comparisonb}, the proposed RGB and RF fusion model demonstrates significant superiority across three key metrics: MAE, RMSE, and $\rho$. Specifically, in terms of MAE (measuring absolute error), our model achieves a minimal skin tone difference of 0.26, which is 61.2\% lower than the suboptimal Vilesov et al. \cite{vilesov2022blending} (0.67) and outperforms conventional methods ICA (4.42) \cite{poh2010advancements}, CHROM (4.97) \cite{dehaan2013robust}, and PhysNet (2.22)~\cite{yu2019remote} by factors of approximately 17, 19, and 8.5, respectively.

Regarding RMSE, our model shows a notable improvement with a value of 1.28, achieving 11.1\% and 68.4\% reductions compared to Vilesov et al. (1.44)~\cite{vilesov2022blending} (1.44) and PhysNet (4.05)~\cite{yu2019remote}, respectively. In terms of $\rho$
, which represents the difference in correlation between light and dark skin samples, our model achieves a minimal difference of 0.05, demonstrating a much smaller skin tone bias compared to the significantly higher differences found in ICA (-0.36) \cite{poh2010advancements}, CHROM (-0.38) \cite{dehaan2013robust}, and PhysNet (-0.25)~\cite{yu2019remote}. This indicates that our approach ensures more consistent and fair performance across different skin tones. These results demonstrate that by integrating multimodal signals from RGB and radio frequency (RF), our model effectively mitigates spectral reflectance estimation biases caused by skin tone variations in traditional unimodal methods, providing a more reliable technical pathway for fairness-sensitive applications.

\begin{figure}[t] 
\centering
\captionsetup{justification=centering}
\begin{subfigure}[t]{0.24\textwidth} 
    \centering
    \includegraphics[width=\linewidth, height=4cm]{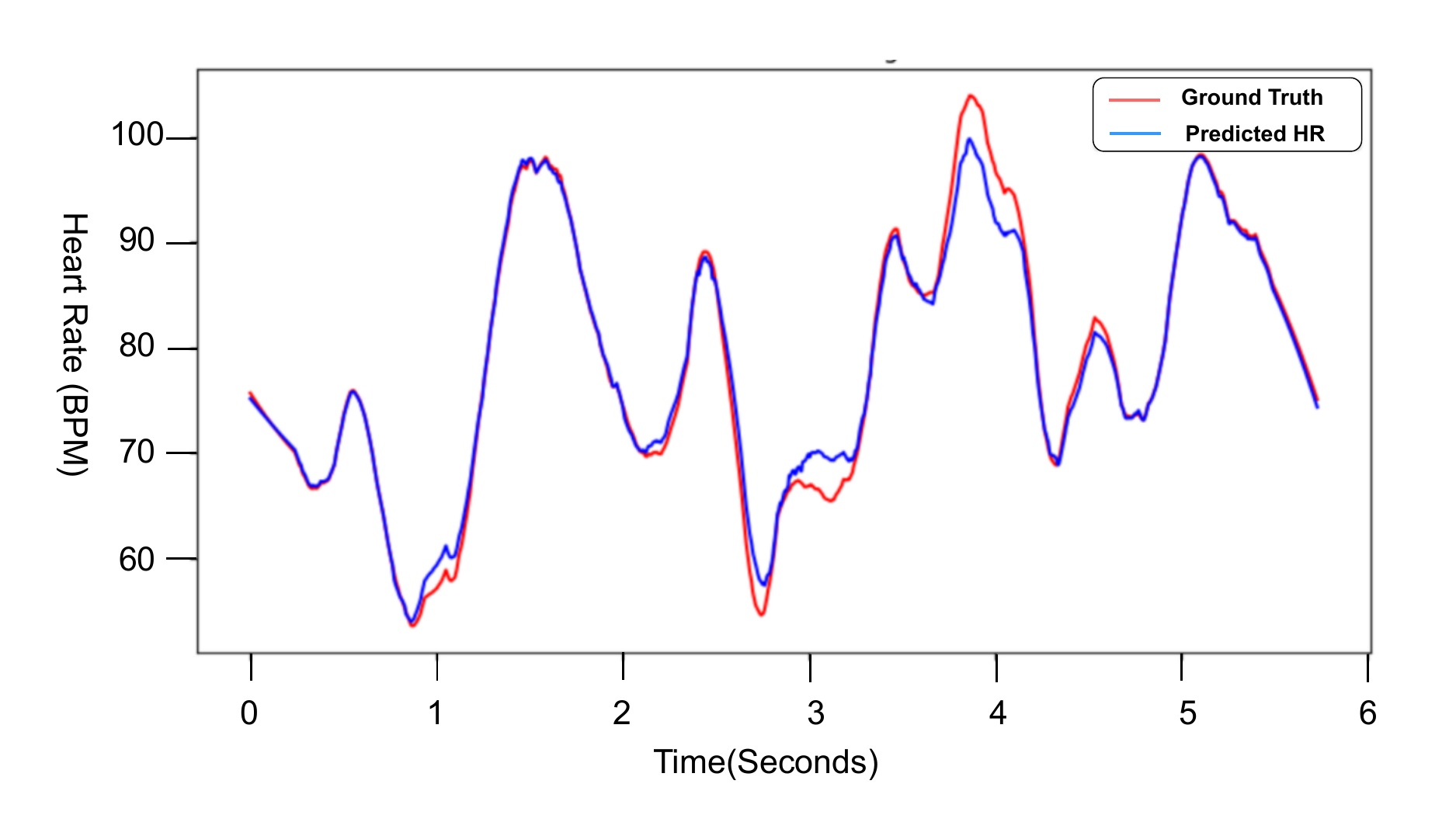}
    \caption{Heart Rate} 
    \label{fig:face_raw3}
\end{subfigure}
\hfill
\begin{subfigure}[t]{0.24\textwidth}
    \centering
    \includegraphics[width=\linewidth, height=3.8cm]{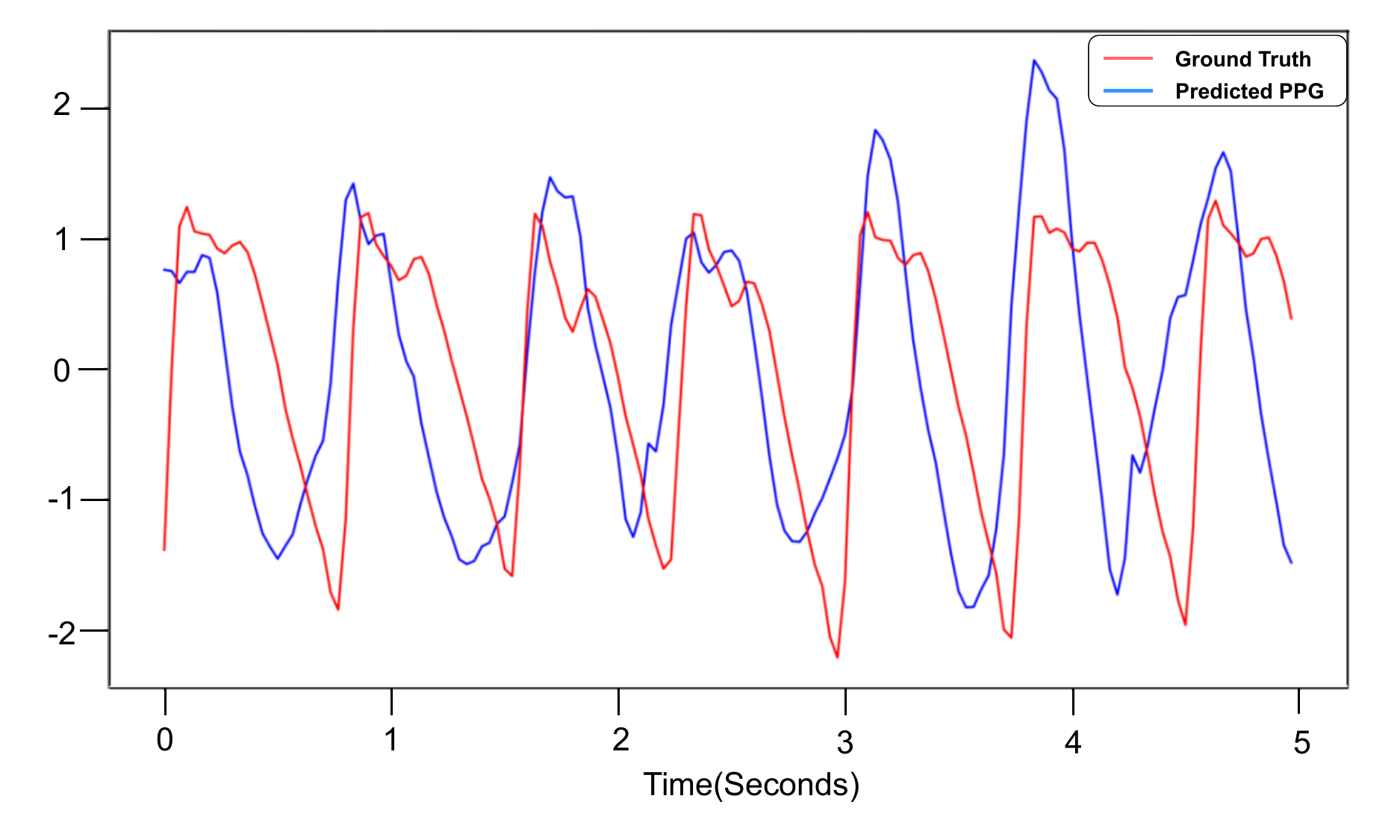}
    \caption{PPG} 
    \label{fig:face_af3}
\end{subfigure}

\caption{ Comparison of ground truth and predicted continuous heart rate and PPG signals. (a) The ground truth (red) and predicted heart rate (blue) over time (in seconds). (b) The ground truth (red) and predicted PPG signals (blue) over time (in seconds). }
\label{fig:full_page_feature3}
\vspace{-0.5em} 
\end{figure}
\subsection{Measurement in Missing Modality Scenarios}
In robustness testings under missing modalities, the proposed multimodal fusion model (RGB and RF) demonstrates exceptional adaptability. As shown in TABLE \ref{tab:comparison3}, the model maintains leading performance in most scenarios even when partial modalities are missing during testing (e.g., RGB-only or RF-only). Under full-modality testing (RGB and RF), our model achieves optimal results outperform Vilesov et al.\cite{vilesov2022blending}  by 14.3\% and 10.5\%, respectively, and surpassing the baseline model by over 20-fold, thereby validating the efficacy of deep multimodal fusion. 

In RGB-only testing, our model reduces MAE by 82.4\% compared to Vilesov et al.~\cite{vilesov2022blending}, with error levels approaching full-modality performance, highlighting the robust generalization of RGB feature extraction. While the model underperforms Vilesov et al.~\cite{vilesov2022blending} (MAE=7.25, RMSE=9.62) in RF-only testing , its superiority under full modalities confirms that RF signals primarily serve as complementary components within the framework, maximizing value through multimodal synergy.

 This capability not only underscores the model’s robustness to modality incompleteness but also ensures reliable deployment in real-world complex environments.

\begin{figure}[t]
\centering
\includegraphics[scale=0.30]{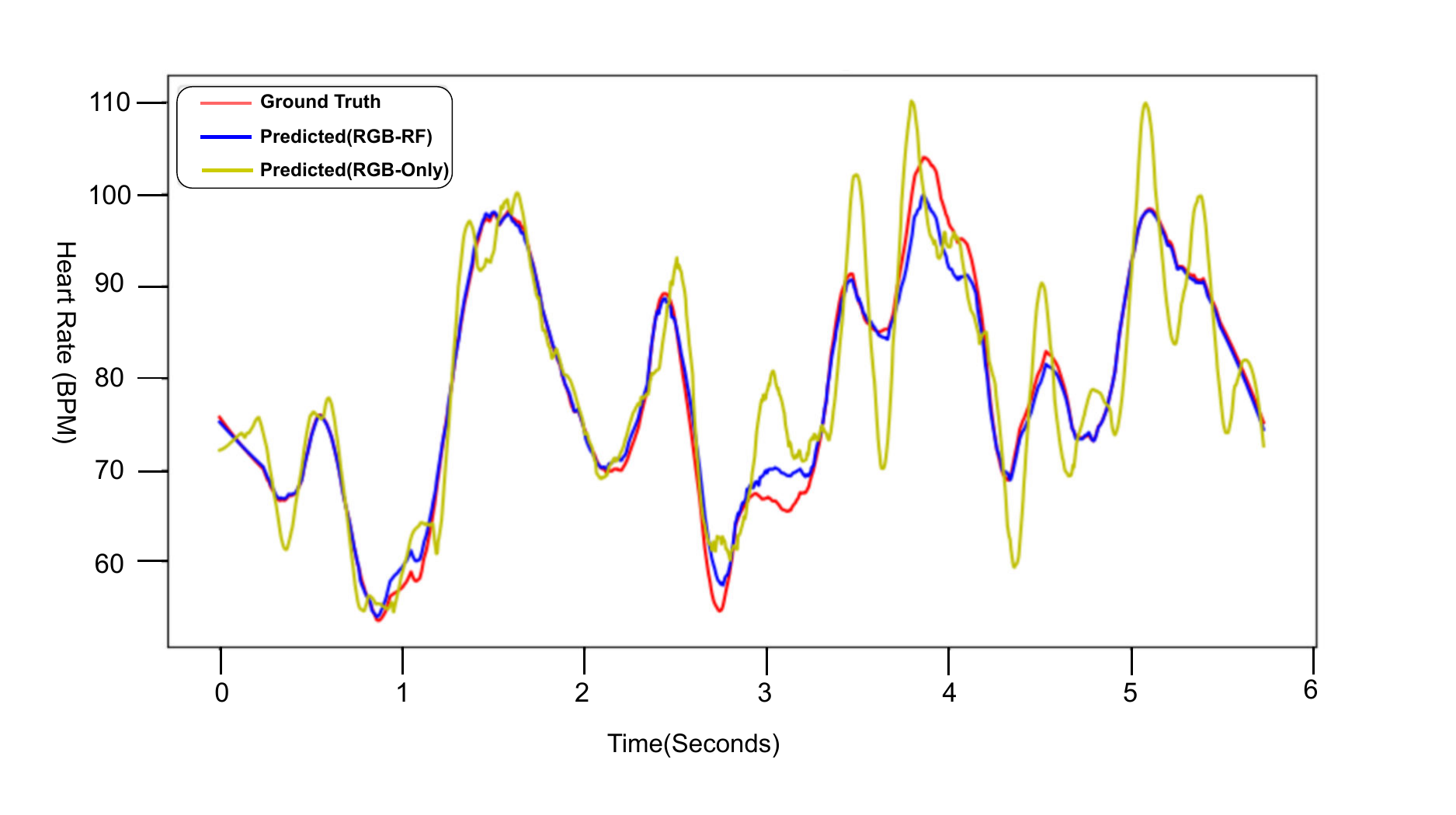}
\vspace{-2.0em}
\caption{Visualization of continuous HR results between RGB-only and RGB-RF fusion Methods  }
\label{fig:4}
\vspace{-1.0em}
\end{figure}

\subsection{Ablation Study}

As shown in TABLE~\ref{tab:comparison4}, to validate the contribution of each module, we conducted systematic ablation studies. The results demonstrate that the Vision Mamba and Channel-wise Fast Fourier Transform (CFFT) are critical to model accuracy. While excluding Vision Mamba (first row) retains decent performance (MAE=1.7, R=0.91), removing CFFT (second row) causes severe degradation (MAE=18.8, $\rho$=0.2), indicating that cross-modal feature alignment heavily relies on frequency-domain compensation. The spatiotemporal synchronization module (SSM) and RF Alignment Module (RFAM) provide auxiliary stabilization and detail enhancement. Their individual absence only slightly increases MAE to 1.862 and 1.85, respectively, while maintaining $\rho$ $>$ 0.89, proving the model’s tolerance to partial module failure. Notably, the exclusion of the Temporal Difference Mamba Module(TDMM)  doubles MAE (3.82 vs. 1.7) and increases RMSE by 64.5\%, highlighting its essential role in temporal consistency regulation. In conclusion, Vision Mamba and CFFT form the core framework, while SSM, channel attention, and Time difference collaboratively refine performance through multi-level optimization, achieving balanced metrics (MAE=1.7, $\rho$=0.91), thereby validating the completeness and robustness of our modules.

\subsection{Visualization and Analysis}
\textbf{Face and Radar Spectrum Feature Visualizations}: As shown in Fig.~\ref{fig:full_page_feature1}, we present the visual representations of both the RGB and RF modalities. Fig.~\ref{fig:face_raw1} shows an image from the EquiPleth dataset, where the human face is captured. Fig.~\ref{fig:face_af2} highlights a feature heatmap of this image, illustrating the key regions of interest for heart rate estimation. The heatmap emphasizes areas with strong signal correlation, such as the face's skin regions. Moving to Fig.~\ref{fig:rf_raw3}, the radar spectrum diagram visualizes the frequency information captured by the radar. Finally, Fig.~\ref{fig:rf_af4} presents the feature heatmap of the radar spectrum, where key frequency components related to physiological signals are emphasized. This series of images provides a clear comparison of the spatial and frequency-domain features from the two modalities.

\textbf{Bland-Altman Plots for CardiacMamba and Other Methods.}: Fig.~\ref{fig:full_page_feature2} shows the Bland-Altman plots comparing the heart rate estimates from CardiacMamba and Vilesov et al. \cite{vilesov2022blending} with the ground truth values. Fig.~\ref{fig:face_raw2} illustrates the Bland-Altman plot for our CardiacMamba model, demonstrating that the heart rate estimates are more tightly clustered within the confidence interval, signifying greater accuracy and consistency. In contrast, Fig.~\ref{fig:face_af2} presents the results from Vilesov et al. \cite{vilesov2022blending}, which show a broader spread and higher variance in heart rate predictions. This comparison highlights the superior performance of CardiacMamba in terms of estimation stability and accuracy.

\textbf{Comparison of Ground Truth and Predicted Signals for Heart Rate and PPG}: As shown in Fig.~\ref{fig:full_page_feature3}, Fig.~\ref{fig:face_af2} compares the ground truth (red) and predicted heart rate (blue) signals over time (in seconds). The predicted heart rate closely follows the ground truth, with minimal fluctuations, indicating high prediction accuracy. The consistency between the two curves demonstrates the robustness of the model in estimating heart rate. Fig.~\ref{fig:face_af3} presents the comparison between the ground truth (red) and predicted PPG signals (blue) over time. The predicted PPG signal accurately mirrors the ground truth, confirming the model's capability in tracking the periodic patterns of the PPG signal. This comparison underscores the model's effectiveness in predicting the physiological signal

\textbf{ Comparison of HR Results Between RGB-only and RGB-RF Fusion Methods}: Fig.~\ref{fig:4} compares the heart rate predictions using two methods: RGB-only (yellow) and RGB-RF Fusion (blue), against the ground truth (red) over time (in seconds). It is clear that the predictions from the RGB-RF fusion model (blue) are much closer to the ground truth, with less fluctuation and greater stability compared to the RGB-only predictions (yellow). The RGB-RF fusion method outperforms the RGB-only approach, demonstrating its superior accuracy, stability, and robustness. The fusion model provides smoother and more consistent heart rate estimation.

\section{Conclusion}
\label{sec:conclusion}
In this paper, we propose the CardiacMamba, a multimodal fusion framework for rPPG heart rate estimation from RGB video and radio frequency (RF) sensors. CardiacMamba achieves dual-level feature extraction and alignment through time-differential perception and convolution alignment modules, enhancing the dynamic features of both RGB and RF modalities, while leveraging Mamba blocks to improve feature expression. Based on a bidirectional state-space model, it performs cross-modal collaborative modeling, preserving the semantic information of each modality and enhancing global context awareness. Additionally, the channel Fourier transform adaptively enhances heart rate-related frequency bands, suppresses noise, and reconstructs temporal features, improving heart rate signal detection. Experimental results demonstrate that this framework outperforms existing methods across multiple performance metrics, effectively mitigates skin tone bias, improves accuracy for darker skin samples, and maintains strong adaptability even in cases of missing modalities, significantly enhancing accuracy, robustness, and fairness.

\ifCLASSOPTIONcaptionsoff
  \newpage
\fi

\bibliographystyle{IEEEtran}
\bibliography{IEEEabrv,reference}

\begin{thebibliography}{10}
\providecommand{\url}[1]{#1}
\csname url@samestyle\endcsname
\providecommand{\newblock}{\relax}
\providecommand{\bibinfo}[2]{#2}
\providecommand{\BIBentrySTDinterwordspacing}{\spaceskip=0pt\relax}
\providecommand{\BIBentryALTinterwordstretchfactor}{4}
\providecommand{\BIBentryALTinterwordspacing}{\spaceskip=\fontdimen2\font plus
\BIBentryALTinterwordstretchfactor\fontdimen3\font minus \fontdimen4\font\relax}
\providecommand{\BIBforeignlanguage}[2]{{%
\expandafter\ifx\csname l@#1\endcsname\relax
\typeout{** WARNING: IEEEtran.bst: No hyphenation pattern has been}%
\typeout{** loaded for the language `#1'. Using the pattern for}%
\typeout{** the default language instead.}%
\else
\language=\csname l@#1\endcsname
\fi
#2}}
\providecommand{\BIBdecl}{\relax}
\BIBdecl

\bibitem{ahad2021contactless}
M.~A.~R. Ahad, U.~Mahbub, and T.~Rahman, \emph{Contactless Human Activity Analysis}.\hskip 1em plus 0.5em minus 0.4em\relax Springer, 2021.

\bibitem{cheng2020remote}
J.~Cheng, P.~Wang, R.~Song, Y.~Liu, C.~Li, Y.~Liu, and X.~Chen, ``Remote heart rate measurement from near-infrared videos based on joint blind source separation with delay-coordinate transformation,'' \emph{IEEE Transactions on Instrumentation and Measurement}, vol.~70, pp. 1--13, 2020.

\bibitem{liu2024remote}
B.~Liu, X.~Zheng, and Y.~I. Wu, ``Remote heart rate estimation in intense interference scenarios: A white-box framework,'' \emph{IEEE Transactions on Instrumentation and Measurement}, 2024.

\bibitem{hu2021eta}
M.~Hu, F.~Qian, D.~Guo, X.~Wang, L.~He, and F.~Ren, ``Eta-rppgnet: Effective time-domain attention network for remote heart rate measurement,'' \emph{IEEE Transactions on Instrumentation and Measurement}, vol.~70, pp. 1--12, 2021.

\bibitem{yu2021facial}
Z.~Yu, X.~Li, and G.~Zhao, ``Facial-video-based physiological signal measurement: Recent advances and affective applications,'' \emph{IEEE Signal Processing Magazine}, vol.~38, no.~6, pp. 50--58, 2021.

\bibitem{dehaan2013robust}
G.~D. Haan and V.~Jeanne, ``Robust pulse rate from chrominance-based rppg,'' \emph{IEEE Transactions on Biomedical Engineering}, vol.~60, no.~10, pp. 2878--2886, 2013.

\bibitem{verkruysse2008remote}
W.~Verkruysse, L.~O. Svaasand, and J.~S. Nelson, ``Remote plethysmographic imaging using ambient light,'' \emph{Optics Express}, vol.~16, no.~26, pp. 21\,434--21\,445, 2008.

\bibitem{lewandowska2011measuring}
M.~Lewandowska, J.~Rumiński, T.~Kocejko, and J.~Nowak, ``Measuring pulse rate with a webcam—a non-contact method for evaluating cardiac activity,'' in \emph{2011 Federated Conference on Computer Science and Information Systems (FedCSIS)}.\hskip 1em plus 0.5em minus 0.4em\relax IEEE, 2011, pp. 405--410.

\bibitem{poh2010advancements}
M.-Z. Poh, D.~J. McDuff, and R.~W. Picard, ``Advancements in noncontact, multiparameter physiological measurements using a webcam,'' \emph{IEEE Transactions on Biomedical Engineering}, vol.~58, no.~1, pp. 7--11, 2010.

\bibitem{yu2019remote}
Z.~Yu, X.~Li, and G.~Zhao, ``Remote photoplethysmograph signal measurement from facial videos using spatio-temporal networks,'' in \emph{The British Machine Vision Conference (BMVC)}, 2019.

\bibitem{nowara2018sparseppg}
E.~M. Nowara, T.~K. Marks, H.~Mansour, and A.~Veeraraghavan, ``Sparseppg: Towards driver monitoring using camera-based vital signs estimation in near-infrared,'' in \emph{Proceedings of the IEEE Conference on Computer Vision and Pattern Recognition Workshops}.\hskip 1em plus 0.5em minus 0.4em\relax IEEE, 2018, pp. 1272--1281.

\bibitem{park2019noncontact}
J.-K. Park, Y.~Hong, H.~Lee, C.~Jang, G.-H. Yun, H.-J. Lee, and J.-G. Yook, ``Noncontact rf vital sign sensor for continuous monitoring of driver status,'' \emph{IEEE Transactions on Biomedical Circuits and Systems}, vol.~13, no.~3, pp. 493--502, 2019.

\bibitem{zheng2020v2ifi}
T.~Zheng, Z.~Chen, C.~Cai, J.~Luo, and X.~Zhang, ``V2ifi: In-vehicle vital sign monitoring via compact rf sensing,'' \emph{Proceedings of the ACM on Interactive, Mobile, Wearable and Ubiquitous Technologies (IMWUT)}, vol.~4, no.~2, pp. 1--27, 2020.

\bibitem{alizadeh2019remote}
M.~Alizadeh, G.~Shaker, J.~C. M.~D. Almeida, P.~P. Morita, and S.~Safavi-Naeini, ``Remote monitoring of human vital signs using mm-wave fmcw radar,'' \emph{IEEE Access}, vol.~7, pp. 54\,958--54\,968, 2019.

\bibitem{li2008random}
C.~Li and J.~Lin, ``Random body movement cancellation in doppler radar vital sign detection,'' \emph{IEEE Transactions on Microwave Theory and Techniques}, vol.~56, no.~12, pp. 3143--3152, 2008.

\bibitem{wu2019person}
S.~Wu, T.~Sakamoto, K.~Oishi, T.~Sato, K.~Inoue, T.~Fukuda, K.~Mizutani, and H.~Sakai, ``Person-specific heart rate estimation with ultra-wideband radar using convolutional neural networks,'' \emph{IEEE Access}, vol.~7, pp. 168\,484--168\,494, 2019.

\bibitem{vilesov2022blending}
A.~Vilesov, P.~Chari, A.~Armouti, A.~B. Harish, K.~Kulkarni, A.~Deoghare, L.~Jalilian, and A.~Kadambi, ``Blending camera and 77 ghz radar sensing for equitable, robust plethysmography,'' \emph{ACM Transactions on Graphics}, vol.~41, no.~4, pp. 1--14, 2022.

\bibitem{balakrishnan2013detecting}
G.~Balakrishnan, F.~Durand, and J.~Guttag, ``Detecting pulse from head motions in video,'' in \emph{Proceedings of the IEEE Conference on Computer Vision and Pattern Recognition (CVPR)}.\hskip 1em plus 0.5em minus 0.4em\relax Portland, OR, USA: IEEE, 2013, pp. 3430--3437.

\bibitem{poh2010noncontact}
M.-Z. Poh, D.~J. McDuff, and R.~W. Picard, ``Noncontact, automated cardiac pulse measurements using video imaging and blind source separation,'' \emph{Optics Express}, vol.~18, no.~10, pp. 10\,762--10\,774, 2010.

\bibitem{monkaresi2014machine}
H.~Monkaresi, R.~A. Calvo, and H.~Yan, ``A machine learning approach to improve contactless heart rate monitoring using a webcam,'' \emph{IEEE Journal of Biomedical and Health Informatics}, vol.~18, no.~4, pp. 1153--1160, 2014.

\bibitem{wang2019vision}
Z.-K. Wang, Y.~Kao, and C.-T. Hsu, ``Vision-based heart rate estimation via a two-stream cnn,'' in \emph{Proceedings of the IEEE International Conference on Image Processing (ICIP)}.\hskip 1em plus 0.5em minus 0.4em\relax Taipei, Taiwan: IEEE, 2019, pp. 3327--3331.

\bibitem{chen2018deepphys}
W.~Chen and D.~McDuff, ``Deepphys: Video-based physiological measurement using convolutional attention networks,'' in \emph{Proceedings of the European Conference on Computer Vision (ECCV)}, 2018, pp. 349--365.

\bibitem{yu2022physformer}
Z.~Yu, Y.~Shen, J.~Shi, H.~Zhao, P.~H. Torr, and G.~Zhao, ``Physformer: Facial video-based physiological measurement with temporal difference transformer,'' in \emph{Proceedings of the IEEE/CVF Conference on Computer Vision and Pattern Recognition}, 2022, pp. 4186--4196.

\bibitem{zou2024rhythmformer}
B.~Zou, Z.~Guo, J.~Chen, and H.~Ma, ``Rhythmformer: Extracting rppg signals based on hierarchical temporal periodic transformer,'' \emph{arXiv preprint arXiv:2402.12788}, 2024.

\bibitem{yu2023physformer++}
Z.~Yu, Y.~Shen, J.~Shi, H.~Zhao, Y.~Cui, J.~Zhang, P.~Torr, and G.~Zhao, ``Physformer++: Facial video-based physiological measurement with slowfast temporal difference transformer,'' \emph{International Journal of Computer Vision}, vol. 131, no.~6, pp. 1307--1330, 2023.

\bibitem{nowara2021benefit}
E.~M. Nowara, D.~McDuff, and A.~Veeraraghavan, ``The benefit of distraction: Denoising camera-based physiological measurements using inverse attention,'' in \emph{Proceedings of the IEEE International Conference on Computer Vision (ICCV)}, 2021, pp. 4955--4964.

\bibitem{lin1975noninvasive}
J.~C. Lin, ``Noninvasive microwave measurement of respiration,'' \emph{Proceedings of the IEEE}, vol.~63, no.~10, pp. 1530--1530, 1975.

\bibitem{ha2020contactless}
U.~Ha, S.~Assana, and F.~Adib, ``Contactless seismocardiography via deep learning radars,'' in \emph{Proceedings of the ACM Annual International Conference on Mobile Computing and Networking (MobiCom)}, 2020, pp. 1--14.

\bibitem{zheng2021morefi}
T.~Zheng, Z.~Chen, S.~Zhang, C.~Cai, and J.~Luo, ``More-fi: Motion-robust and fine-grained respiration monitoring via deep-learning uwb radar,'' in \emph{Proceedings of the ACM Conference on Embedded Networked Sensor Systems (SenSys)}, New York, NY, USA, 2021, pp. 111--124.

\bibitem{negishi2020contactless}
T.~Negishi, S.~Abe, T.~Matsui, H.~Liu, M.~Kurosawa, T.~Kirimoto, and G.~Sun, ``Contactless vital signs measurement system using rgb-thermal image sensors and its clinical screening test on patients with seasonal influenza,'' \emph{Sensors}, vol.~20, no.~8, p. 2171, 2020.

\bibitem{matsumura2020rgb}
K.~Matsumura, S.~Toda, and Y.~Kato, ``Rgb and near-infrared light reflectance/transmittance photoplethysmography for measuring heart rate during motion,'' \emph{IEEE Access}, vol.~8, pp. 80\,233--80\,242, 2020.

\bibitem{park2022self}
S.~Park, B.-K. Kim, and S.-Y. Dong, ``Self-supervised rgb-nir fusion video vision transformer framework for rppg estimation,'' \emph{IEEE Transactions on Instrumentation and Measurement}, vol.~71, pp. 1--10, 2022.

\bibitem{gu2022efficiently}
A.~Gu, K.~Goel, and C.~Ré, ``Efficiently modeling long sequences with structured state spaces,'' in \emph{Proceedings of the Tenth International Conference on Learning Representations (ICLR)}, 2022.

\bibitem{fu2023hungry}
D.~Y. Fu, T.~Dao, K.~K. Saab, A.~W. Thomas, A.~Rudra, and C.~Ré, ``Hungry hungry hippos: Towards language modeling with state space models,'' in \emph{Proceedings of the Eleventh International Conference on Learning Representations (ICLR)}, 2023.

\bibitem{mehta2023long}
H.~Mehta, A.~Gupta, A.~Cutkosky, and B.~Neyshabur, ``Long range language modeling via gated state spaces,'' in \emph{Proceedings of the Eleventh International Conference on Learning Representations (ICLR 2023)}, 2023.

\bibitem{smith2023simplified}
J.~T.~H. Smith, A.~Warrington, and S.~W. Linderman, ``Simplified state space layers for sequence modeling,'' in \emph{Proceedings of the Eleventh International Conference on Learning Representations (ICLR)}, 2023.

\bibitem{xie2024fusionmamba}
X.~Xie, Y.~Cui, T.~Tan, X.~Zheng, and Z.~Yu, ``Fusionmamba: Dynamic feature enhancement for multimodal image fusion with mamba,'' \emph{Visual Intelligence}, vol.~2, no.~1, p.~37, 2024.

\bibitem{luo2024physmamba}
C.~Luo, Y.~Xie, and Z.~Yu, ``Physmamba: Efficient remote physiological measurement with slowfast temporal difference mamba,'' in \emph{Chinese Conference on Biometric Recognition}, 2024, pp. 248--259.

\bibitem{zhu2024vision}
L.~Zhu, B.~Liao, Q.~Zhang, X.~Wang, W.~Liu, and X.~Wang, ``Vision mamba: Efficient visual representation learning with bidirectional state space model,'' in \emph{Proceedings of the International Conference on Machine Learning (ICML)}, 2024.

\bibitem{liu2020multi}
X.~Liu, J.~Fromm, S.~Patel, and D.~McDuff, ``Multi-task temporal shift attention networks for on-device contactless vitals measurement,'' in \emph{Advances in Neural Information Processing Systems (NeurIPS)}, Virtual, 2020, pp. 1--23.

\bibitem{liu2023efficientphys}
X.~Liu, B.~Hill, Z.~Jiang, S.~Patel, and D.~McDuff, ``Efficientphys: Enabling simple, fast and accurate camera-based cardiac measurement,'' in \emph{Proceedings of the IEEE/CVF Winter Conference on Applications of Computer Vision}, 2023, pp. 5008--5017.

\bibitem{zou2024rhythmmamba}
B.~Zou, Z.~Guo, X.~Hu, and H.~Ma, ``Rhythmmamba: Fast remote physiological measurement with arbitrary length videos,'' \emph{arXiv preprint arXiv:2404.06483}, 2024.

\bibitem{tu2016respiration}
J.~Tu, T.~Hwang, and J.~Lin, ``Respiration rate measurement under 1-d body motion using single continuous-wave doppler radar vital sign detection system,'' \emph{IEEE Transactions on Microwave Theory and Techniques}, vol.~64, no.~6, pp. 1937--1946, 2016.

\bibitem{mercuri2019vital}
M.~Mercuri, I.~Lorato, Y.-H. Liu, P.~Wieringa, F.~V. Hoof, and T.~Torfs, ``Vital-sign monitoring and spatial tracking of multiple people using a contactless radar-based sensor,'' \emph{Nature Electronics}, vol.~2, pp. 252--262, 2019.

\bibitem{ye2024depmamba}
\BIBentryALTinterwordspacing
J.~Ye, J.~Zhang, and H.~Shan, ``Depmamba: Progressive fusion mamba for multimodal depression detection,'' \emph{arXiv preprint arXiv:2409.15936}, 2024. [Online]. Available: \url{https://arxiv.org/abs/2409.15936}
\BIBentrySTDinterwordspacing

\bibitem{zhang2016joint}
K.~Zhang, Z.~Zhang, Z.~Li, and Y.~Qiao, ``Joint face detection and alignment using multitask cascaded convolutional networks,'' \emph{IEEE Signal Processing Letters}, vol.~23, no.~10, pp. 1499--1503, 2016.

\end{thebibliography}


\end{document}